%% file: main.tex
\documentclass[sigconf,nonacm]{acmart}

\usepackage{amsmath}
\usepackage{booktabs}
\usepackage{multirow}
\usepackage{threeparttable}
\usepackage{subcaption}
\usepackage{balance}
\usepackage{array}
\usepackage{colortbl}
\usepackage{enumitem}
\usepackage{tikz}
\usepackage[skins,breakable]{tcolorbox}
\usetikzlibrary{arrows.meta,backgrounds,calc,fit,positioning}
\input{package.tex}
\input{styles/components.tex}

\input{styles/vendor/lucide/lucide_icons.tex}
\input{styles/figures.tex}

\title{Proof-of-Agentness: Towards Verifying Agents on the Web}
\author{Zuyao Xu}
\email{xuzuyao@mail.nankai.edu.cn}
\affiliation{\institution{Nankai University}\city{Tianjin}\country{China}}
\author{Xiang Li}
\authornote{Corresponding author.}
\email{lixiang@nankai.edu.cn}
\affiliation{\institution{Nankai University}\city{Tianjin}\country{China}}

\begin{document}
\PaperPageLayout
\input{sections/00_abstract.tex}

\maketitle

\input{sections/01_introduction.tex}

\input{sections/02_background.tex}
\input{sections/03_problem.tex}
\input{sections/04_design.tex}

\input{sections/05_evaluation.tex}

\input{sections/06_discussion.tex}
\input{sections/07_conclusion.tex}

\clearpage
\bibliographystyle{styles/vendor/acm/ACM-Reference-Format}
\bibliography{references}

\appendix
\setcounter{dbltopnumber}{3}
\input{appendix/a_challenge_details.tex}
\input{appendix/b_passive_details.tex}

\end{document}

%% file: package.tex
\setcopyright{none}
\DeclareRobustCommand{\PoA}{\textit{PoA}}
\DeclareRobustCommand{\AgentScore}{\textsc{AgentScore}}

\newcommand{\PoAModelLogoSize}{7.2pt}
\newcommand{\PoAModelLogoDrop}{1.2pt}
\newcommand{\PoAModelLogoGap}{2.4pt}
\newcommand{\PoAPreviewTableColumnSep}{3pt}
\newcommand{\PoAPreviewTableRowStretch}{1.05}
\newcommand{\PoAAccessTableColumnSep}{1.2pt}
\newcommand{\PoATableRoleIconSize}{2.5mm}
\newcommand{\PoATaskIconSize}{2.1ex}
\newcommand{\PoATaskIconGap}{.25em}

\newcommand{\PaperPageLayout}{\raggedbottom}
\AtBeginDocument{\setcounter{dbltopnumber}{2}}
\newcommand{\PoAFloatFont}{\small}

\definecolor{PoABackground}{HTML}{FFFFFF}
\definecolor{PoAFg}{HTML}{0A0A0A}
\definecolor{PoAPrimary}{HTML}{171717}
\definecolor{PoAMuted}{HTML}{F5F5F5}
\definecolor{PoAMutedFg}{HTML}{737373}
\definecolor{PoAChallengeTitle}{HTML}{505050}
\definecolor{PoABorder}{HTML}{E5E5E5}
\definecolor{PoARing}{HTML}{A3A3A3}
\definecolor{PoAAccent}{HTML}{B91C1C}
\definecolor{PoAAccentSoft}{HTML}{FEF2F2}
\definecolor{PoAAccentBorder}{HTML}{B77979}
\definecolor{PoAPlotText}{HTML}{4A342B}
\definecolor{PoABot}{HTML}{B8860B}
\colorlet{PoAOnDark}{PoABackground}
\colorlet{PoAOnDarkMuted}{PoABorder}
\newcommand{\PoAFigureWidth}{17.5cm}
\newcommand{\PoAComparisonWidth}{\PoAFigureWidth}
\newcommand{\PoAFigureRadius}{3pt}
\newcommand{\PoAFigureLineWidth}{0.55pt}
\newcommand{\PoAFigureArrowWidth}{0.65pt}
\newcommand{\PoAFigureFontFamily}{\fontfamily{phv}}
\newcommand{\PoAFigureFont}{\PoAFigureFontFamily\fontsize{7}{9.2}\selectfont}
\newcommand{\PoAFigureSmallFont}{\PoAFigureFontFamily\fontsize{6.6}{8.8}\selectfont}

\newcommand{\PoAFigureSymbolFont}{\PoAFigureFontFamily\fontsize{8.5}{10.5}\selectfont}
\newcommand{\PoAFigureTitleFont}{\PoAFigureFontFamily\bfseries\fontsize{7.6}{9.6}\selectfont}

\graphicspath{{assets/images/}}

\newcommand\xl[1]{{\color{blue}{\begin{CJK}{UTF8}{gbsn}\textbf{\{xiang: {\em#1}\}}\end{CJK}}}}

\newcommand{\ignore}[1]{}
\newcommand{\ignoredbyxiang}[1]{}

%% file: styles/components.tex
\newcommand{\PoAFloatStyle}{\centering\PoAFloatFont}
\newcommand{\PoAModelName}[2]{%
  \raisebox{-\PoAModelLogoDrop}{#1}\hspace{\PoAModelLogoGap}#2%
}
\newcommand{\PoATaskName}[2]{%
  \raisebox{-.45ex}{\includegraphics[height=\PoATaskIconSize]{assets/icons/tasks/#1.pdf}}\nobreak\hspace{\PoATaskIconGap}#2%
}
\newcommand{\PoAScriptName}[2]{%
  \PoAModelName{\includegraphics[width=\PoAModelLogoSize]{assets/icons/lucide/#1.pdf}}{\textsc{#2}}%
}
\newcommand{\PoAPreviewTableStyle}{%
  \PoAFloatStyle
  \setlength{\tabcolsep}{\PoAPreviewTableColumnSep}%
  \renewcommand{\arraystretch}{\PoAPreviewTableRowStretch}%
}
\newcommand{\PoAClientIcons}[3]{%
  \ifnum#1=1{\color{PoAFg}\PoAIcon[\PoATableRoleIconSize]{user-round}}\fi%
  \ifnum#2=1%
    \ifnum#1=1\hspace{\PoAPreviewTableColumnSep}\fi%
    {\color{PoABot}\PoAIcon[\PoATableRoleIconSize]{bot}}%
  \fi%
  \ifnum#3=1%
    \ifnum\numexpr#1+#2\relax>0\relax\hspace{\PoAPreviewTableColumnSep}\fi%
    {\color{PoAAccent}\PoAIcon[\PoATableRoleIconSize]{brain-circuit}}%
  \fi%
}
\newcommand{\PoAAccessEntry}[2]{%
  \textbf{#1}: #2%
}
\newtcolorbox{poachallenge}{%
  enhanced,
  breakable,
  colback=PoABackground,
  colframe=PoAMutedFg,
  colbacktitle=PoAChallengeTitle,
  coltitle=PoAOnDark,
  fonttitle=\bfseries,
  title={Public challenge},
  title after break={Public challenge (continued)},
  boxrule=0.45pt,
  arc=2pt,
  boxsep=0pt,
  left=6pt,
  right=6pt,
  top=5pt,
  bottom=5pt,
  toptitle=4pt,
  bottomtitle=4pt,
  before skip=7pt,
  after skip=7pt,
  lines before break=12,
  before upper={\setlength{\parindent}{0pt}\setlength{\parskip}{3pt}},
}
\newtcolorbox{poadefinition}[1][]{%
  enhanced,
  colback=PoAMuted,
  colframe=PoAPrimary,
  colbacktitle=PoAMuted,
  coltitle=PoAFg,
  fonttitle=\bfseries,
  boxrule=0.45pt,
  arc=2pt,
  borderline west={2pt}{0pt}{PoAPrimary},
  boxsep=0pt,
  left=8pt,
  right=8pt,
  top=7pt,
  bottom=6pt,
  toptitle=6pt,
  bottomtitle=3pt,
  titlerule=0pt,
  before skip=8pt,
  after skip=8pt,
  #1
}

%% file: styles/vendor/lucide/lucide_icons.tex
\usetikzlibrary{svg.path}

\newdimen\PoAIconSize
\newdimen\PoAIconStroke

\newcommand{\PoAIcon}[2][4mm]{%
  \begingroup
  \PoAIconSize=#1\relax
  \PoAIconStroke=\PoAIconSize
  \divide\PoAIconStroke by 12
  \colorlet{PoAIconCurrent}{.}%
  \begin{tikzpicture}[baseline=.15\PoAIconSize,sharp corners]
    \pgfpathrectangle{\pgfpointorigin}{\pgfpoint{\PoAIconSize}{\PoAIconSize}}%
    \pgfusepath{use as bounding box}%
    \pgfsetlinewidth{\PoAIconStroke}%
    \pgfsetdash{}{0pt}%
    \pgfsetarrows{-}%
    \pgfsetroundcap
    \pgfsetroundjoin
    \pgfsetstrokecolor{PoAIconCurrent}%
    \pgfmathsetmacro{\PoAIconScale}{\the\PoAIconSize/24pt}%
    \pgftransformshift{\pgfpoint{0pt}{\PoAIconSize}}%
    \pgftransformscale{\PoAIconScale}%
    \pgftransformyscale{-1}%
    \ifcsname PoALucide@#2\endcsname
      \csname PoALucide@#2\endcsname
    \else
      \PackageError{poa-lucide}{Unknown Lucide icon '#2'}{See styles/vendor/lucide/README.md for the supported icon names.}%
    \fi
  \end{tikzpicture}%
  \endgroup
}

\expandafter\def\csname PoALucide@activity\endcsname{%
  \pgfpathsvg{M 22 12 L 19.52 12 C 18.62181268 11.99807874 17.83250908 12.59516851 17.59 13.46 L 15.24 21.82 C 15.20888889 21.92666667 15.11111111 22 15 22 C 14.88888889 22 14.79111111 21.92666667 14.76 21.82 L 9.24 2.18 C 9.208888889 2.073333333 9.111111111 2 9 2 C 8.888888889 2 8.791111111 2.073333333 8.76 2.18 L 6.41 10.54 C 6.16849394 11.40126076 5.38447679 11.99744047 4.49 12 L 2 12}%
  \pgfusepath{stroke}%
}

\expandafter\def\csname PoALucide@bot\endcsname{%
  \pgfpathsvg{M 12 8 L 12 4 L 8 4}%
  \pgfusepath{stroke}%
  \pgfpathsvg{M 6 8 L 18 8 C 19.1045695 8 20 8.8954305 20 10 L 20 18 C 20 19.1045695 19.1045695 20 18 20 L 6 20 C 4.8954305 20 4 19.1045695 4 18 L 4 10 C 4 8.8954305 4.8954305 8 6 8 Z}%
  \pgfusepath{stroke}%
  \pgfpathsvg{M 2 14 L 4 14}%
  \pgfusepath{stroke}%
  \pgfpathsvg{M 20 14 L 22 14}%
  \pgfusepath{stroke}%
  \pgfpathsvg{M 15 13 L 15 15}%
  \pgfusepath{stroke}%
  \pgfpathsvg{M 9 13 L 9 15}%
  \pgfusepath{stroke}%
}

\expandafter\def\csname PoALucide@brain-circuit\endcsname{%
  \pgfpathsvg{M 12 5 C 12.00320809 3.911193126 11.41623229 2.906119945 10.46634338 2.37393411 C 9.516454474 1.841748274 8.352785975 1.866003496 7.425897958 2.43730841 C 6.499009941 3.008613324 5.954424757 4.037272485 6.003 5.125 C 4.811979238 5.431239448 3.8280214 6.268309966 3.334843595 7.39484834 C 2.841665791 8.521386714 2.894089969 9.81216624 3.477 10.895 C 2.450026675 11.72931898 1.902089844 13.01751033 2.013363823 14.3359869 C 2.124637802 15.65446348 2.880722589 16.83260078 4.033 17.483 C 3.842013092 18.96066709 4.48815103 20.42201954 5.709695028 21.27513517 C 6.931239027 22.1282508 8.525710506 22.23172033 9.847267277 21.54363333 C 11.16882405 20.85554634 11.99840979 19.48995754 12 18 Z}%
  \pgfusepath{stroke}%
  \pgfpathsvg{M 9 13 C 10.71688542 12.39602524 11.90091215 10.81732294 12 9}%
  \pgfusepath{stroke}%
  \pgfpathsvg{M 6.003 5.125 C 6.022775001 5.608733518 6.159328622 6.080495397 6.401 6.5}%
  \pgfusepath{stroke}%
  \pgfpathsvg{M 3.477 10.896 C 3.659939503 10.74700544 3.855702739 10.61448878 4.062 10.5}%
  \pgfusepath{stroke}%
  \pgfpathsvg{M 6 18 C 5.310829955 18.0003217 4.6332671 17.82257771 4.033 17.484}%
  \pgfusepath{stroke}%
  \pgfpathsvg{M 12 13 L 16 13}%
  \pgfusepath{stroke}%
  \pgfpathsvg{M 12 18 L 18 18 C 19.1045695 18 20 18.8954305 20 20 L 20 21}%
  \pgfusepath{stroke}%
  \pgfpathsvg{M 12 8 L 20 8}%
  \pgfusepath{stroke}%
  \pgfpathsvg{M 16 8 L 16 5 C 16 3.8954305 16.8954305 3 18 3}%
  \pgfusepath{stroke}%
  \pgfpathsvg{M 16.5 13 C 16.5 12.72385763 16.27614237 12.5 16 12.5 C 15.72385763 12.5 15.5 12.72385763 15.5 13 C 15.5 13.27614237 15.72385763 13.5 16 13.5 C 16.27614237 13.5 16.5 13.27614237 16.5 13 Z}%
  \pgfusepath{stroke}%
  \pgfpathsvg{M 18.5 3 C 18.5 2.723857625 18.27614237 2.5 18 2.5 C 17.72385763 2.5 17.5 2.723857625 17.5 3 C 17.5 3.276142375 17.72385763 3.5 18 3.5 C 18.27614237 3.5 18.5 3.276142375 18.5 3 Z}%
  \pgfusepath{stroke}%
  \pgfpathsvg{M 20.5 21 C 20.5 20.72385763 20.27614237 20.5 20 20.5 C 19.72385763 20.5 19.5 20.72385763 19.5 21 C 19.5 21.27614237 19.72385763 21.5 20 21.5 C 20.27614237 21.5 20.5 21.27614237 20.5 21 Z}%
  \pgfusepath{stroke}%
  \pgfpathsvg{M 20.5 8 C 20.5 7.723857625 20.27614237 7.5 20 7.5 C 19.72385763 7.5 19.5 7.723857625 19.5 8 C 19.5 8.276142375 19.72385763 8.5 20 8.5 C 20.27614237 8.5 20.5 8.276142375 20.5 8 Z}%
  \pgfusepath{stroke}%
}

\expandafter\def\csname PoALucide@circle-question-mark\endcsname{%
  \pgfpathsvg{M 22 12 C 22 6.477152502 17.5228475 2 12 2 C 6.477152502 2 2 6.477152502 2 12 C 2 17.5228475 6.477152502 22 12 22 C 17.5228475 22 22 17.5228475 22 12 Z}%
  \pgfusepath{stroke}%
  \pgfpathsvg{M 9.09 9 C 9.575436876 7.620044439 10.98538362 6.791406324 12.42717631 7.038712103 C 13.868969 7.286017882 14.92217936 8.537152926 14.92 10 C 14.92 12 11.92 13 11.92 13}%
  \pgfusepath{stroke}%
  \pgfpathsvg{M 12 17 L 12.01 17}%
  \pgfusepath{stroke}%
}

\expandafter\def\csname PoALucide@clock-3\endcsname{%
  \pgfpathsvg{M 22 12 C 22 6.477152502 17.5228475 2 12 2 C 6.477152502 2 2 6.477152502 2 12 C 2 17.5228475 6.477152502 22 12 22 C 17.5228475 22 22 17.5228475 22 12 Z}%
  \pgfusepath{stroke}%
  \pgfpathsvg{M 12 6 L 12 12 L 16 12}%
  \pgfusepath{stroke}%
}

\expandafter\def\csname PoALucide@cpu\endcsname{%
  \pgfpathsvg{M 12 20 L 12 22}%
  \pgfusepath{stroke}%
  \pgfpathsvg{M 12 2 L 12 4}%
  \pgfusepath{stroke}%
  \pgfpathsvg{M 17 20 L 17 22}%
  \pgfusepath{stroke}%
  \pgfpathsvg{M 17 2 L 17 4}%
  \pgfusepath{stroke}%
  \pgfpathsvg{M 2 12 L 4 12}%
  \pgfusepath{stroke}%
  \pgfpathsvg{M 2 17 L 4 17}%
  \pgfusepath{stroke}%
  \pgfpathsvg{M 2 7 L 4 7}%
  \pgfusepath{stroke}%
  \pgfpathsvg{M 20 12 L 22 12}%
  \pgfusepath{stroke}%
  \pgfpathsvg{M 20 17 L 22 17}%
  \pgfusepath{stroke}%
  \pgfpathsvg{M 20 7 L 22 7}%
  \pgfusepath{stroke}%
  \pgfpathsvg{M 7 20 L 7 22}%
  \pgfusepath{stroke}%
  \pgfpathsvg{M 7 2 L 7 4}%
  \pgfusepath{stroke}%
  \pgfpathsvg{M 6 4 L 18 4 C 19.1045695 4 20 4.8954305 20 6 L 20 18 C 20 19.1045695 19.1045695 20 18 20 L 6 20 C 4.8954305 20 4 19.1045695 4 18 L 4 6 C 4 4.8954305 4.8954305 4 6 4 Z}%
  \pgfusepath{stroke}%
  \pgfpathsvg{M 9 8 L 15 8 C 15.55228475 8 16 8.44771525 16 9 L 16 15 C 16 15.55228475 15.55228475 16 15 16 L 9 16 C 8.44771525 16 8 15.55228475 8 15 L 8 9 C 8 8.44771525 8.44771525 8 9 8 Z}%
  \pgfusepath{stroke}%
}

\expandafter\def\csname PoALucide@database\endcsname{%
  \pgfpathsvg{M 21 5 C 21 3.343145751 16.97056275 2 12 2 C 7.029437252 2 3 3.343145751 3 5 C 3 6.656854249 7.029437252 8 12 8 C 16.97056275 8 21 6.656854249 21 5 Z}%
  \pgfusepath{stroke}%
  \pgfpathsvg{M 3 5 L 3 19 C 3 20.65685425 7.029437252 22 12 22 C 16.97056275 22 21 20.65685425 21 19 L 21 5}%
  \pgfusepath{stroke}%
  \pgfpathsvg{M 3 12 C 3 13.65685425 7.029437252 15 12 15 C 16.97056275 15 21 13.65685425 21 12}%
  \pgfusepath{stroke}%
}

\expandafter\def\csname PoALucide@file-text\endcsname{%
  \pgfpathsvg{M 6 22 C 4.8954305 22 4 21.1045695 4 20 L 4 4 C 4 2.8954305 4.8954305 2 6 2 L 14 2 C 14.63936507 1.998963747 15.25271507 2.253086459 15.704 2.706 L 19.292 6.294 C 19.74613988 6.745444061 20.00104038 7.359653461 20 8 L 20 20 C 20 21.1045695 19.1045695 22 18 22 Z}%
  \pgfusepath{stroke}%
  \pgfpathsvg{M 14 2 L 14 7 C 14 7.55228475 14.44771525 8 15 8 L 20 8}%
  \pgfusepath{stroke}%
  \pgfpathsvg{M 10 9 L 8 9}%
  \pgfusepath{stroke}%
  \pgfpathsvg{M 16 13 L 8 13}%
  \pgfusepath{stroke}%
  \pgfpathsvg{M 16 17 L 8 17}%
  \pgfusepath{stroke}%
}

\expandafter\def\csname PoALucide@fingerprint-pattern\endcsname{%
  \pgfpathsvg{M 12 10 C 10.8954305 10 10 10.8954305 10 12 C 10 13.02 9.9 14.51 9.74 16}%
  \pgfusepath{stroke}%
  \pgfpathsvg{M 14 13.12 C 14 15.5 14 19.5 13 22}%
  \pgfusepath{stroke}%
  \pgfpathsvg{M 17.29 21.02 C 17.41 20.42 17.72 18.72 17.79 18}%
  \pgfusepath{stroke}%
  \pgfpathsvg{M 2 12 C 2 7.695695919 4.754300949 3.874307483 8.83772234 2.513167019 C 12.92114373 1.152026556 17.41741755 2.556556735 20 6}%
  \pgfusepath{stroke}%
  \pgfpathsvg{M 2 16 L 2.01 16}%
  \pgfusepath{stroke}%
  \pgfpathsvg{M 21.8 16 C 22 14 21.931 10.646 21.8 10}%
  \pgfusepath{stroke}%
  \pgfpathsvg{M 5 19.5 C 5.5 18 6 15 6 12 C 5.998989525 11.31886729 6.113965989 10.64253515 6.34 10}%
  \pgfusepath{stroke}%
  \pgfpathsvg{M 8.65 22 C 8.86 21.34 9.1 20.68 9.22 20}%
  \pgfusepath{stroke}%
  \pgfpathsvg{M 9 6.8 C 10.85700177 5.727858939 13.14500859 5.728225403 15.00166682 6.800961271 C 16.85832505 7.873697138 18.0013753 9.855719201 18 12 L 18 14}%
  \pgfusepath{stroke}%
}

\expandafter\def\csname PoALucide@gauge\endcsname{%
  \pgfpathsvg{M 12 14 L 16 10}%
  \pgfusepath{stroke}%
  \pgfpathsvg{M 3.34 19 C 0.9132543173 14.7972463 1.854384053 9.458777015 5.571994174 6.339224446 C 9.289604294 3.219671877 14.71039571 3.219671877 18.42800583 6.339224446 C 22.14561595 9.458777015 23.08674568 14.7972463 20.66 19}%
  \pgfusepath{stroke}%
}

\expandafter\def\csname PoALucide@key-round\endcsname{%
  \pgfpathsvg{M 2.586 17.414 C 2.21090125 17.78898546 2.000113275 18.29761037 2 18.828 L 2 21 C 2 21.55228475 2.44771525 22 3 22 L 6 22 C 6.55228475 22 7 21.55228475 7 21 L 7 20 C 7 19.44771525 7.44771525 19 8 19 L 9 19 C 9.55228475 19 10 18.55228475 10 18 L 10 17 C 10 16.44771525 10.44771525 16 11 16 L 11.172 16 C 11.70238963 15.99988673 12.21101454 15.78909875 12.586 15.414 L 13.4 14.6 C 16.26221079 15.5970426 19.43361793 14.48358621 21.04435913 11.91612494 C 22.65510032 9.348663668 22.27759917 6.008738332 20.13443042 3.865569583 C 17.99126167 1.722400833 14.65133633 1.344899678 12.08387506 2.955640873 C 9.516413789 4.566382068 8.402957398 7.737789206 9.4 10.6 Z}%
  \pgfusepath{stroke}%
  \pgfpathsvg{M 17 7.5 C 17 7.223857625 16.77614237 7 16.5 7 C 16.22385763 7 16 7.223857625 16 7.5 C 16 7.776142375 16.22385763 8 16.5 8 C 16.77614237 8 17 7.776142375 17 7.5 Z}%
  \pgfusepath{stroke}%
}

\expandafter\def\csname PoALucide@list-ordered\endcsname{%
  \pgfpathsvg{M 11 5 L 21 5}%
  \pgfusepath{stroke}%
  \pgfpathsvg{M 11 12 L 21 12}%
  \pgfusepath{stroke}%
  \pgfpathsvg{M 11 19 L 21 19}%
  \pgfusepath{stroke}%
  \pgfpathsvg{M 4 4 L 5 4 L 5 9}%
  \pgfusepath{stroke}%
  \pgfpathsvg{M 4 9 L 6 9}%
  \pgfusepath{stroke}%
  \pgfpathsvg{M 6.5 20 L 3.4 20 C 3.4 19 6 18.075 6 16.5 C 6.000080676 15.88293259 5.622257621 15.3287744 5.047813845 15.10341569 C 4.473370069 14.87805698 3.81952517 15.02748389 3.4 15.48}%
  \pgfusepath{stroke}%
}

\expandafter\def\csname PoALucide@lock-keyhole\endcsname{%
  \pgfpathsvg{M 13 16 C 13 15.44771525 12.55228475 15 12 15 C 11.44771525 15 11 15.44771525 11 16 C 11 16.55228475 11.44771525 17 12 17 C 12.55228475 17 13 16.55228475 13 16 Z}%
  \pgfusepath{stroke}%
  \pgfpathsvg{M 5 10 L 19 10 C 20.1045695 10 21 10.8954305 21 12 L 21 20 C 21 21.1045695 20.1045695 22 19 22 L 5 22 C 3.8954305 22 3 21.1045695 3 20 L 3 12 C 3 10.8954305 3.8954305 10 5 10 Z}%
  \pgfusepath{stroke}%
  \pgfpathsvg{M 7 10 L 7 7 C 7 4.238576251 9.238576251 2 12 2 C 14.76142375 2 17 4.238576251 17 7 L 17 10}%
  \pgfusepath{stroke}%
}

\expandafter\def\csname PoALucide@messages-square\endcsname{%
  \pgfpathsvg{M 16 10 C 16 11.1045695 15.1045695 12 14 12 L 6.828 12 C 6.297610373 12.00011327 5.788985463 12.21090125 5.414 12.586 L 3.212 14.788 C 3.008943026 14.99101992 2.70359175 15.05174659 2.438307155 14.94186798 C 2.17302256 14.83198938 2.000037053 14.57313973 2 14.286 L 2 4 C 2 2.8954305 2.8954305 2 4 2 L 14 2 C 15.1045695 2 16 2.8954305 16 4 Z}%
  \pgfusepath{stroke}%
  \pgfpathsvg{M 20 9 C 21.1045695 9 22 9.8954305 22 11 L 22 21.286 C 21.99996295 21.57313973 21.82697744 21.83198938 21.56169284 21.94186798 C 21.29640825 22.05174659 20.99105697 21.99101992 20.788 21.788 L 18.586 19.586 C 18.21101454 19.21090125 17.70238963 19.00011327 17.172 19 L 10 19 C 8.8954305 19 8 18.1045695 8 17 L 8 16}%
  \pgfusepath{stroke}%
}

\expandafter\def\csname PoALucide@repeat-2\endcsname{%
  \pgfpathsvg{M 2 9 L 5 6 L 8 9}%
  \pgfusepath{stroke}%
  \pgfpathsvg{M 13 18 L 7 18 C 5.8954305 18 5 17.1045695 5 16 L 5 6}%
  \pgfusepath{stroke}%
  \pgfpathsvg{M 22 15 L 19 18 L 16 15}%
  \pgfusepath{stroke}%
  \pgfpathsvg{M 11 6 L 17 6 C 18.1045695 6 19 6.8954305 19 8 L 19 18}%
  \pgfusepath{stroke}%
}

\expandafter\def\csname PoALucide@route\endcsname{%
  \pgfpathsvg{M 9 19 C 9 17.34314575 7.656854249 16 6 16 C 4.343145751 16 3 17.34314575 3 19 C 3 20.65685425 4.343145751 22 6 22 C 7.656854249 22 9 20.65685425 9 19 Z}%
  \pgfusepath{stroke}%
  \pgfpathsvg{M 9 19 L 17.5 19 C 19.43299662 19 21 17.43299662 21 15.5 C 21 13.56700338 19.43299662 12 17.5 12 L 6.5 12 C 4.567003376 12 3 10.43299662 3 8.5 C 3 6.567003376 4.567003376 5 6.5 5 L 15 5}%
  \pgfusepath{stroke}%
  \pgfpathsvg{M 21 5 C 21 3.343145751 19.65685425 2 18 2 C 16.34314575 2 15 3.343145751 15 5 C 15 6.656854249 16.34314575 8 18 8 C 19.65685425 8 21 6.656854249 21 5 Z}%
  \pgfusepath{stroke}%
}

\expandafter\def\csname PoALucide@scan-eye\endcsname{%
  \pgfpathsvg{M 3 7 L 3 5 C 3 3.8954305 3.8954305 3 5 3 L 7 3}%
  \pgfusepath{stroke}%
  \pgfpathsvg{M 17 3 L 19 3 C 20.1045695 3 21 3.8954305 21 5 L 21 7}%
  \pgfusepath{stroke}%
  \pgfpathsvg{M 21 17 L 21 19 C 21 20.1045695 20.1045695 21 19 21 L 17 21}%
  \pgfusepath{stroke}%
  \pgfpathsvg{M 7 21 L 5 21 C 3.8954305 21 3 20.1045695 3 19 L 3 17}%
  \pgfusepath{stroke}%
  \pgfpathsvg{M 13 12 C 13 11.44771525 12.55228475 11 12 11 C 11.44771525 11 11 11.44771525 11 12 C 11 12.55228475 11.44771525 13 12 13 C 12.55228475 13 13 12.55228475 13 12 Z}%
  \pgfusepath{stroke}%
  \pgfpathsvg{M 18.944 12.33 C 19.01869209 12.11633967 19.01869209 11.88366033 18.944 11.67 C 17.79241075 8.84820034 15.04773869 7.003877908 12 7.003877908 C 8.952261312 7.003877908 6.207589245 8.84820034 5.056 11.67 C 4.981307909 11.88366033 4.981307909 12.11633967 5.056 12.33 C 6.207589245 15.15179966 8.952261312 16.99612209 12 16.99612209 C 15.04773869 16.99612209 17.79241075 15.15179966 18.944 12.33}%
  \pgfusepath{stroke}%
}

\expandafter\def\csname PoALucide@server\endcsname{%
  \pgfpathsvg{M 4 2 L 20 2 C 21.1045695 2 22 2.8954305 22 4 L 22 8 C 22 9.1045695 21.1045695 10 20 10 L 4 10 C 2.8954305 10 2 9.1045695 2 8 L 2 4 C 2 2.8954305 2.8954305 2 4 2 Z}%
  \pgfusepath{stroke}%
  \pgfpathsvg{M 4 14 L 20 14 C 21.1045695 14 22 14.8954305 22 16 L 22 20 C 22 21.1045695 21.1045695 22 20 22 L 4 22 C 2.8954305 22 2 21.1045695 2 20 L 2 16 C 2 14.8954305 2.8954305 14 4 14 Z}%
  \pgfusepath{stroke}%
  \pgfpathsvg{M 6 6 L 6.01 6}%
  \pgfusepath{stroke}%
  \pgfpathsvg{M 6 18 L 6.01 18}%
  \pgfusepath{stroke}%
}

\expandafter\def\csname PoALucide@shield-check\endcsname{%
  \pgfpathsvg{M 20 13 C 20 18 16.5 20.5 12.34 21.95 C 12.12216248 22.02381746 11.88553747 22.02028575 11.67 21.94 C 7.5 20.5 4 18 4 13 L 4 6 C 4 5.44771525 4.44771525 5 5 5 C 7 5 9.5 3.8 11.24 2.28 C 11.67767373 1.906067264 12.32232627 1.906067264 12.76 2.28 C 14.51 3.81 17 5 19 5 C 19.55228475 5 20 5.44771525 20 6 Z}%
  \pgfusepath{stroke}%
  \pgfpathsvg{M 9 12 L 11 14 L 15 10}%
  \pgfusepath{stroke}%
}

\expandafter\def\csname PoALucide@shield-question-mark\endcsname{%
  \pgfpathsvg{M 20 13 C 20 18 16.5 20.5 12.34 21.95 C 12.12216248 22.02381746 11.88553747 22.02028575 11.67 21.94 C 7.5 20.5 4 18 4 13 L 4 6 C 4 5.44771525 4.44771525 5 5 5 C 7 5 9.5 3.8 11.24 2.28 C 11.67767373 1.906067264 12.32232627 1.906067264 12.76 2.28 C 14.51 3.81 17 5 19 5 C 19.55228475 5 20 5.44771525 20 6 Z}%
  \pgfusepath{stroke}%
  \pgfpathsvg{M 9.1 9 C 9.595281445 7.635030641 10.99703204 6.820711731 12.42810973 7.066601368 C 13.85918742 7.312491004 14.90873475 8.547995126 14.92 10 C 14.92 12 11.92 13 11.92 13}%
  \pgfusepath{stroke}%
  \pgfpathsvg{M 12 17 L 12.01 17}%
  \pgfusepath{stroke}%
}

\expandafter\def\csname PoALucide@sliders-horizontal\endcsname{%
  \pgfpathsvg{M 10 5 L 3 5}%
  \pgfusepath{stroke}%
  \pgfpathsvg{M 12 19 L 3 19}%
  \pgfusepath{stroke}%
  \pgfpathsvg{M 14 3 L 14 7}%
  \pgfusepath{stroke}%
  \pgfpathsvg{M 16 17 L 16 21}%
  \pgfusepath{stroke}%
  \pgfpathsvg{M 21 12 L 12 12}%
  \pgfusepath{stroke}%
  \pgfpathsvg{M 21 19 L 16 19}%
  \pgfusepath{stroke}%
  \pgfpathsvg{M 21 5 L 14 5}%
  \pgfusepath{stroke}%
  \pgfpathsvg{M 8 10 L 8 14}%
  \pgfusepath{stroke}%
  \pgfpathsvg{M 8 12 L 3 12}%
  \pgfusepath{stroke}%
}

\expandafter\def\csname PoALucide@terminal\endcsname{%
  \pgfpathsvg{M 12 19 L 20 19}%
  \pgfusepath{stroke}%
  \pgfpathsvg{M 4 17 L 10 11 L 4 5}%
  \pgfusepath{stroke}%
}

\expandafter\def\csname PoALucide@ticket-check\endcsname{%
  \pgfpathsvg{M 2 9 C 3.656854249 9 5 10.34314575 5 12 C 5 13.65685425 3.656854249 15 2 15 L 2 17 C 2 18.1045695 2.8954305 19 4 19 L 20 19 C 21.1045695 19 22 18.1045695 22 17 L 22 15 C 20.34314575 15 19 13.65685425 19 12 C 19 10.34314575 20.34314575 9 22 9 L 22 7 C 22 5.8954305 21.1045695 5 20 5 L 4 5 C 2.8954305 5 2 5.8954305 2 7 Z}%
  \pgfusepath{stroke}%
  \pgfpathsvg{M 9 12 L 11 14 L 15 10}%
  \pgfusepath{stroke}%
}

\expandafter\def\csname PoALucide@trash\endcsname{%
  \pgfpathsvg{M 10 11 L 10 17}%
  \pgfusepath{stroke}%
  \pgfpathsvg{M 14 11 L 14 17}%
  \pgfusepath{stroke}%
  \pgfpathsvg{M 19 6 L 19 20 C 19 21.1045695 18.1045695 22 17 22 L 7 22 C 5.8954305 22 5 21.1045695 5 20 L 5 6}%
  \pgfusepath{stroke}%
  \pgfpathsvg{M 3 6 L 21 6}%
  \pgfusepath{stroke}%
  \pgfpathsvg{M 8 6 L 8 4 C 8 2.8954305 8.8954305 2 10 2 L 14 2 C 15.1045695 2 16 2.8954305 16 4 L 16 6}%
  \pgfusepath{stroke}%
}

\expandafter\def\csname PoALucide@user-round\endcsname{%
  \pgfpathsvg{M 17 8 C 17 5.238576251 14.76142375 3 12 3 C 9.238576251 3 7 5.238576251 7 8 C 7 10.76142375 9.238576251 13 12 13 C 14.76142375 13 17 10.76142375 17 8 Z}%
  \pgfusepath{stroke}%
  \pgfpathsvg{M 20 21 C 20 16.581722 16.418278 13 12 13 C 7.581722001 13 4 16.581722 4 21}%
  \pgfusepath{stroke}%
}

\expandafter\def\csname PoALucide@workflow\endcsname{%
  \pgfpathsvg{M 5 3 L 9 3 C 10.1045695 3 11 3.8954305 11 5 L 11 9 C 11 10.1045695 10.1045695 11 9 11 L 5 11 C 3.8954305 11 3 10.1045695 3 9 L 3 5 C 3 3.8954305 3.8954305 3 5 3 Z}%
  \pgfusepath{stroke}%
  \pgfpathsvg{M 7 11 L 7 15 C 7 16.1045695 7.8954305 17 9 17 L 13 17}%
  \pgfusepath{stroke}%
  \pgfpathsvg{M 15 13 L 19 13 C 20.1045695 13 21 13.8954305 21 15 L 21 19 C 21 20.1045695 20.1045695 21 19 21 L 15 21 C 13.8954305 21 13 20.1045695 13 19 L 13 15 C 13 13.8954305 13.8954305 13 15 13 Z}%
  \pgfusepath{stroke}%
}

\expandafter\def\csname PoALucide@circle-help\endcsname{\csname PoALucide@circle-question-mark\endcsname}

\expandafter\def\csname PoALucide@shield-question\endcsname{\csname PoALucide@shield-question-mark\endcsname}

\expandafter\def\csname PoALucide@fingerprint\endcsname{\csname PoALucide@fingerprint-pattern\endcsname}

%% file: styles/figures.tex
\tikzset{
  poa/picture/.style={
    x=1cm, y=1cm, font=\PoAFigureFont,
    line cap=round, line join=round, text=PoAFg
  },
  poa/card/.style={
    draw=PoABorder, fill=PoABackground,
    line width=\PoAFigureLineWidth,
    rounded corners=\PoAFigureRadius,
    inner sep=4pt, align=left
  },
  poa/panel/.style={poa/card, fill=PoAMuted},
  poa/dark surface/.style={fill=PoAPrimary, draw=PoAPrimary, text=PoAOnDark},
  poa/accent surface/.style={fill=PoAAccent, draw=PoAAccent, text=PoAOnDark},
  poa/accent soft/.style={fill=PoAAccentSoft, draw=PoAAccentBorder},
  poa/inverse text/.style={text=PoAOnDark},
  poa/inverse secondary/.style={text=PoAOnDarkMuted},
  poa/accent text/.style={text=PoAAccent},
  poa/arrow/.style={
    draw=PoAPrimary, line width=\PoAFigureArrowWidth,
    -{Latex[length=1.6mm,width=1.1mm]}, rounded corners=2pt
  },
  poa/dashed arrow/.style={
    poa/arrow, draw=PoAMutedFg, dash pattern=on 2.2pt off 1.7pt
  },
  poa/accent arrow/.style={poa/arrow, draw=PoAAccent},
  poa/accent dashed arrow/.style={poa/dashed arrow, draw=PoAAccent},
  poa/label/.style={font=\PoAFigureFont, text=PoAFg, inner sep=0pt, align=left},
  poa/small/.style={font=\PoAFigureSmallFont, text=PoAFg, inner sep=0pt, align=left},
  poa/title/.style={font=\PoAFigureTitleFont, text=PoAFg, inner sep=0pt, align=left},
  poa/badge/.style={
    draw=PoABorder, fill=PoAMuted, text=PoAPrimary,
    font=\PoAFigureSmallFont, rounded corners=2pt,
    line width=\PoAFigureLineWidth, inner xsep=4pt, inner ysep=2pt
  },
  poa/proposed/.style={
    draw=PoAAccent, dash pattern=on 2.2pt off 1.7pt
  }
}

%% file: sections/00_abstract.tex
\begin{abstract}
CAPTCHAs and bot detection have helped protect Web services from automated abuse for decades by distinguishing humans from bots.
However, the emergence of LLM-powered agents introduces a new setting: Web services increasingly accommodate agents as legitimate users.
Without a way to distinguish agents from other automated clients such as simple bots and scripts, these services risk large-scale automated junk submissions that waste service resources and consume users' attention or tokens.
In this paper, we investigate a novel question for today's Web: \textit{``How can a Web service obtain evidence that a client is an agent?''}
We introduce the concept of \textit{``Proof-of-Agentness''} (\PoA{}), a mechanism for distinguishing agents from other automated clients, motivated by the parallel between human attention and an agent's inference resources.
We outline two approaches: \textit{active verification} through challenge--response and \textit{passive analysis} of service-visible signals.
Eight language models achieve an overall accuracy of 96.5\% on our challenges, with an average overhead of 1,647 tokens.
For passive analysis, we conduct an IRB-approved user study comparing the Web interactions of twelve human participants and representative Bot and Agent implementations, achieving an average classification accuracy of 95.8\% across four classifiers, motivating further study.
\end{abstract}

%% file: sections/01_introduction.tex
\section{Introduction}
\label{sec:introduction}
CAPTCHAs and other bot detection mechanisms have helped protect human-facing Web services from automated abuse at scale by requiring evidence of human participation before granting access~\cite{vonahn2003captcha,park2006robot}.
Each operation performed by a human consumes time and attention, while scripts can cheaply repeat requests and submissions, overwhelming services.
These mechanisms aim to make access depend on evidence of human participation that bots cannot cheaply reproduce.

Meanwhile, Web services are increasingly being designed for agents as well as humans.
Cloudflare and Vercel serve agent-friendly Markdown pages~\cite{cloudflare2026markdownagents,vercel2026agentpages}.
Many online services provide agent access through the Model Context Protocol (MCP), covering search (Brave, Exa, and Tavily)~\cite{bravemcp,examcp,tavilymcp}; data and knowledge queries (Hugging Face, Kaggle, Wikidata)~\cite{huggingfacemcp,kagglemcp,wikidatamcp}; payments (Stripe and PayPal)~\cite{stripe2026mcp,paypal2026mcp}; shopping (Shopify)~\cite{shopifyagents}; and development and workplace collaboration (GitHub, Notion, and Atlassian)~\cite{githubmcp,notionmcp,atlassianmcp}.
However, these interfaces can also be used by ordinary programs.
When agents rely on large language models (LLMs) to interpret tasks and choose actions, they consume inference compute and tokens~\cite{sardana2024inference}.
A traditional automated program can avoid such resource costs and repeatedly submit requests directly, occupying server-side queues and consuming storage and computing resources.
Junk submissions produced in this way can also require users or agents to spend attention or tokens processing their contents.

The resource analogy between human attention and model inference motivates our central research question:
\textit{Can a Web service obtain evidence that a client is an agent} and protect service capacity for useful agent work, just as CAPTCHAs and bot detection protect human-facing services for genuine human interaction?

Agent verification is already entering Web infrastructure.
Cloudflare's Signed agents and Verified bots programs recognize agent traffic, and Web Bot Auth checks signed HTTP requests against registered public keys~\cite{cloudflare2025signedagents,cloudflare2026verifiedbots,cloudflare2026webbotauth}.
However, these approaches authenticate identities asserted by agent operators or infrastructure providers without establishing actual agent participation.
Recent studies also explore browser and behavioral fingerprints to distinguish Web agents from human users and other automated clients~\cite{wang2026fpagent,fayolle2026llmbot}.
These efforts advance agent recognition, but the positioning of agent verification on the Web remains unclear, and the concept itself is not yet fully developed.

\noindent \textbf{Our Study}
In this paper, we introduce \textit{Proof-of-Agentness} (\PoA{}), which asks for evidence of language model participation in an interaction.
We outline two design approaches: \textit{active verification} that issues challenges to clients and checks their responses, and \textit{passive analysis} using Web traffic and interaction characteristics.
We then introduce a unified \AgentScore{} that combines evidence for admission decisions.
Such mechanisms could empower Web services to restrict access to agents acting on users' behalf and regulate automated traffic more selectively, limiting large-scale scripted abuse of service resources.

\begin{figure*}[t]
  \centering
  \input{figures/01_motivation.tex}%
  \caption{Human users, AI agents, and scripted abuse on the Web.
  (a) CAPTCHA seeks human-participation evidence to protect human-facing services from automated abuse.
  (b) Agents and LLM workflows spend tokens interpreting tasks and feedback; scripts can replay junk without such costs while consuming shared service resources.
  Motivated by preserving capacity for useful agent work, \textit{Proof-of-Agentness} asks for evidence of language model participation.}
  \Description{Two panels contrast CAPTCHA-based human verification with PoA's goal of verifying model participation.
  Agents use inference to interpret tasks and feedback, while scripts can repeat requests without inference, occupy the same service's queues, and consume its storage and computing resources.}
  \label{fig:motivation}
\end{figure*}
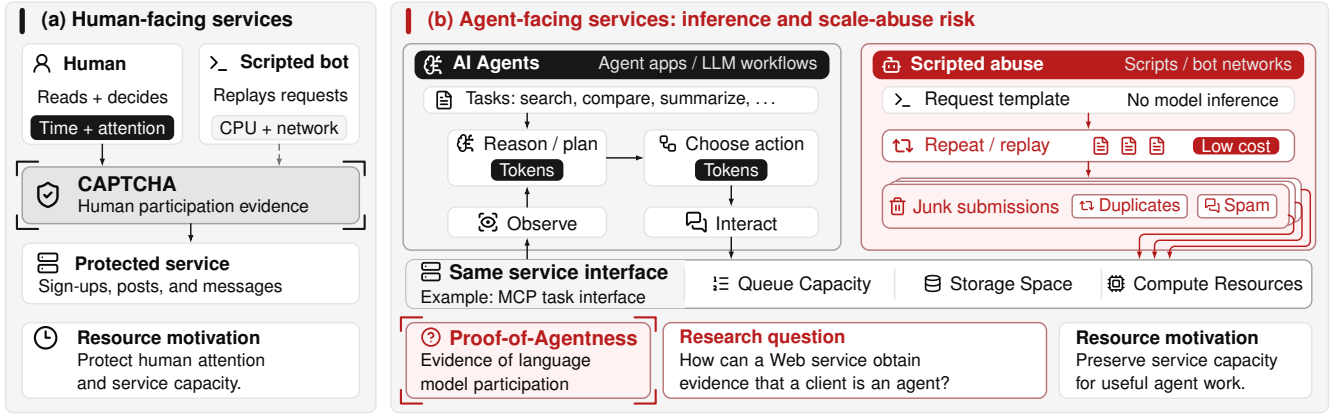

We evaluate active verification with eight language models and two rule-based scripts on 100 challenges, achieving an overall accuracy of 96.5\% with an average overhead of 1,647 tokens.
For passive analysis, we conduct an IRB-approved study of 720 Web visits across three shopping tasks, involving twelve human participants alongside six Bot and three Agent configurations.
Using 40 service-visible features, we achieve an average classification accuracy of 95.8\% across four classifiers on held-out tasks.
Figure~\ref{fig:motivation} connects the CAPTCHA analogy to agent verification and scripted abuse at a shared Web service.

\noindent
\textbf{Contributions.}
We make three contributions:
\begin{itemize}
  \item \textbf{Concept positioning.}
  We introduce \textit{Proof-of-Agentness} (\PoA{}) as a Web verification objective centered on language model participation, analyzing its potential benefits and trade-offs for Web services.
  \item \textbf{Verification design.}
  We outline two approaches to verifying agent participation: active verification through a challenge--response mechanism and passive analysis of Web traffic and interaction characteristics.
  \item \textbf{Empirical study.}
  We evaluate active verification with eight language models, examining challenge performance and overhead.
  To explore passive recognition, we collect a Web traffic and interaction dataset from twelve human participants and representative Bot and Agent implementations.
\end{itemize}

%% file: figures/01_motivation.tex
\newcommand{\PoAHumanPanel}{(a) Human-facing services}%
\newcommand{\PoAHumanClient}{Human}%
\newcommand{\PoAHumanDetail}{Reads + decides}%
\newcommand{\PoAHumanResource}{Time + attention}%
\newcommand{\PoAScriptClient}{Scripted bot}%
\newcommand{\PoAScriptDetail}{Replays requests}%
\newcommand{\PoAScriptCost}{CPU + network}%
\newcommand{\PoAHumanGate}{CAPTCHA}%
\newcommand{\PoAHumanEvidence}{Human participation evidence}%
\newcommand{\PoAHumanService}{Protected service}%
\newcommand{\PoAHumanServiceDetail}{Sign-ups, posts, and messages}%
\newcommand{\PoAHumanMotivation}{Resource motivation}%
\newcommand{\PoAHumanProtect}{Protect human attention\\and service capacity.}%
\newcommand{\PoAModelPanel}{(b) Agent-facing services: inference and scale-abuse risk}%
\newcommand{\PoAModelWorkflow}{AI Agents}%
\newcommand{\PoAModelDetail}{Agent apps / LLM workflows}%
\newcommand{\PoAModelTask}{Tasks: search, compare, summarize, \ldots}%
\newcommand{\PoAModelReason}{Reason / plan}%
\newcommand{\PoAModelChoose}{Choose action}%
\newcommand{\PoAModelTokens}{Tokens}%
\newcommand{\PoAModelInteract}{Interact}%
\newcommand{\PoAModelObserve}{Observe}%
\newcommand{\PoAAbuseHeading}{Scripted abuse}%
\newcommand{\PoAAbuseDetail}{Scripts / bot networks}%
\newcommand{\PoAAbuseTemplate}{Request template}%
\newcommand{\PoAAbuseReplay}{Repeat / replay}%
\newcommand{\PoAAbuseNoInference}{No model inference}%
\newcommand{\PoAAbuseLowCost}{Low cost}%
\newcommand{\PoAAbuseJunk}{Junk submissions}%
\newcommand{\PoAAbuseDuplicates}{Duplicates}%
\newcommand{\PoAAbuseSpam}{Spam}%
\newcommand{\PoAModelService}{Same service interface}%
\newcommand{\PoAModelServiceExample}{Example: MCP task interface}%
\newcommand{\PoAServiceQueue}{Queue Capacity}%
\newcommand{\PoAServiceStorage}{Storage Space}%
\newcommand{\PoAServiceCompute}{Compute Resources}%
\newcommand{\PoAModelGate}{Proof-of-Agentness}%
\newcommand{\PoAModelEvidence}{Evidence of language\\model participation}%
\newcommand{\PoAModelStatus}{Research question}%
\newcommand{\PoAModelQuestion}{How can a Web service obtain\\evidence that a client is an agent?}%
\newcommand{\PoAModelMotivation}{Resource motivation}%
\newcommand{\PoAModelProtect}{Preserve service capacity\\for useful agent work.}%

\begin{tikzpicture}[poa/picture]
  \tikzset{
    poa/comparison focus/.style={line width=.8pt,rounded corners=0pt},
    poa/arrow/.append style={line width=.55pt,-{Latex[length=1.1mm,width=.75mm]}},
    poa/comparison example/.style={
      poa/badge,fill=PoABackground,draw=PoAAccentBorder,text=PoAAccent,
      inner xsep=2.5pt,inner ysep=1.1pt,rounded corners=1.5pt
    }
  }
  \coordinate (comparison-bottom) at (0,-.16);
  \path[use as bounding box] (comparison-bottom) rectangle (\PoAComparisonWidth,5.27);
  \path[poa/panel] (comparison-bottom) rectangle (4.90,5.27);
  \path[poa/panel] ([xshift=5.10cm]comparison-bottom) rectangle (17.50,5.27);
  \draw[draw=PoAPrimary,line width=2pt] (.22,4.88) -- (.22,5.16);
  \node[poa/title,anchor=west] at (.47,5.02) {\PoAHumanPanel};
  \draw[draw=PoAAccent,line width=2pt] (5.33,4.88) -- (5.33,5.16);
  \node[poa/title,poa/accent text,anchor=west] at (5.58,5.02) {\PoAModelPanel};

  \path[poa/card] (.22,3.39) rectangle (2.34,4.72);
  \node[poa/label] at (.48,4.46) {\PoAIcon[3mm]{user-round}};
  \node[poa/label,font=\PoAFigureFont\bfseries,anchor=west] at (.76,4.46) {\PoAHumanClient};
  \node[poa/small,anchor=west] at (.41,4.02) {\PoAHumanDetail};
  \node[poa/badge,poa/dark surface,inner xsep=3pt] at (1.28,3.61) {\PoAHumanResource};
  \path[poa/card] (2.56,3.39) rectangle (4.68,4.72);
  \node[poa/label] at (2.82,4.46) {\PoAIcon[3mm]{terminal}};
  \node[poa/label,font=\PoAFigureFont\bfseries,anchor=west] at (3.10,4.46) {\PoAScriptClient};
  \node[poa/small,anchor=west] at (2.74,4.02) {\PoAScriptDetail};
  \node[poa/badge,inner xsep=3pt,text=PoAFg] at (3.62,3.61) {\PoAScriptCost};
  \draw[poa/arrow] (1.28,3.39) -- (1.28,3.09);
  \draw[poa/dashed arrow] (3.62,3.39) -- (3.62,3.09);
  \path[poa/card,draw=PoARing,fill=PoABorder] (.22,2.34) rectangle (4.68,3.09);
  \foreach \xx/\dx in {.15/.28,4.75/-.28} {
    \draw[poa/comparison focus,draw=PoAPrimary]
      (\xx,2.98) -- (\xx,3.16) -- (\xx+\dx,3.16);
    \draw[poa/comparison focus,draw=PoAPrimary]
      (\xx,2.45) -- (\xx,2.27) -- (\xx+\dx,2.27);}
  \node[poa/label] at (.56,2.72) {\PoAIcon[3.8mm]{shield-check}};
  \node[poa/title,anchor=west] at (.95,2.87) {\PoAHumanGate};
  \node[poa/small,anchor=west] at (.95,2.55) {\PoAHumanEvidence};
  \draw[poa/arrow] (2.45,2.34) -- (2.45,2.08);
  \path[poa/card] (.22,1.28) rectangle (4.68,2.08);
  \node[poa/label] at (.56,1.82) {\PoAIcon[3mm]{server}};
  \node[poa/label,font=\PoAFigureFont\bfseries,anchor=west] at (.91,1.82) {\PoAHumanService};
  \node[poa/small,anchor=west] at (.43,1.50) {\PoAHumanServiceDetail};
  \path[poa/card] (.22,.03) rectangle (4.68,1.03);
  \node[poa/label] at (.53,.83) {\PoAIcon[3.4mm]{clock-3}};
  \node[poa/label,font=\PoAFigureFont\bfseries,anchor=west] at (.94,.83) {\PoAHumanMotivation};
  \node[poa/small,anchor=north west,text width=3.50cm] at (.94,.63) {\PoAHumanProtect};

  \path[poa/panel,draw=PoARing] (5.26,1.99) rectangle (11.05,4.72);
  \path[poa/panel,poa/accent soft] (11.32,1.99) rectangle (17.33,4.72);
  \path[poa/card,poa/dark surface] (5.42,4.26) rectangle (10.89,4.62);
  \path[poa/card,poa/accent surface] (11.48,4.26) rectangle (17.17,4.62);

  \node[poa/label,poa/inverse text] at (5.66,4.44) {\PoAIcon[2.8mm]{brain-circuit}};
  \node[poa/label,font=\PoAFigureFont\bfseries,poa/inverse text,anchor=west] at (5.91,4.44) {\PoAModelWorkflow};
  \node[poa/small,poa/inverse secondary,anchor=east] at (10.75,4.44) {\PoAModelDetail};
  \path[poa/card] (5.53,3.80) rectangle (10.75,4.15);
  \node[poa/label] at (5.80,3.975) {\PoAIcon[2.7mm]{file-text}};
  \node[poa/small,text=PoAFg,anchor=west] at (6.08,3.975) {\PoAModelTask};
  \path[poa/card] (5.85,2.84) rectangle (7.95,3.58);
  \node[poa/label] at (6.90,3.40)
    {\PoAIcon[2.6mm]{brain-circuit}\hspace{3pt}\PoAModelReason};
  \node[poa/badge,poa/dark surface,inner xsep=3pt,inner ysep=1.4pt] at (6.90,3.05) {\PoAModelTokens};
  \path[poa/card] (8.45,2.84) rectangle (10.75,3.58);
  \node[poa/label] at (9.60,3.40)
    {\PoAIcon[2.6mm]{workflow}\hspace{3pt}\PoAModelChoose};
  \node[poa/badge,poa/dark surface,inner xsep=3pt,inner ysep=1.4pt] at (9.60,3.05) {\PoAModelTokens};
  \path[poa/card] (8.45,2.16) rectangle (10.75,2.56);
  \node[poa/label] at (9.60,2.36)
    {\PoAIcon[3mm]{messages-square}\hspace{3pt}\PoAModelInteract};
  \path[poa/card] (5.85,2.16) rectangle (7.95,2.56);
  \node[poa/label] at (6.90,2.36)
    {\PoAIcon[3mm]{scan-eye}\hspace{3pt}\PoAModelObserve};
  \draw[poa/arrow] (6.90,3.80) -- (6.90,3.58);
  \draw[poa/arrow] (7.95,3.21) -- (8.45,3.21);
  \draw[poa/arrow] (9.60,2.84) -- (9.60,2.56);
  \draw[poa/arrow] (9.60,2.16) -- (9.60,1.86);
  \draw[poa/arrow] (6.90,1.86) -- (6.90,2.16);
  \draw[poa/arrow] (6.90,2.56) -- (6.90,2.84);

  \node[poa/label,poa/inverse text] at (11.72,4.44) {\PoAIcon[2.8mm]{bot}};
  \node[poa/label,font=\PoAFigureFont\bfseries,poa/inverse text,anchor=west] at (11.97,4.44) {\PoAAbuseHeading};
  \node[poa/small,poa/inverse secondary,anchor=east] at (17.03,4.44) {\PoAAbuseDetail};
  \path[poa/card] (11.60,3.80) rectangle (17.05,4.15);
  \node[poa/label] at (11.87,3.975) {\PoAIcon[2.7mm]{terminal}};
  \node[poa/label,anchor=west] at (12.15,3.975) {\PoAAbuseTemplate};
  \node[poa/small,text=PoAFg,anchor=east] at (16.85,3.975) {\PoAAbuseNoInference};
  \path[poa/card,draw=PoAAccentBorder] (11.60,3.16) rectangle (17.05,3.58);
  \node[poa/label,poa/accent text] at (11.87,3.37) {\PoAIcon[3mm]{repeat-2}};
  \node[poa/label,poa/accent text,anchor=west] at (12.15,3.37) {\PoAAbuseReplay};
  \foreach \xx in {14.49,14.86,15.23} {
    \node[poa/label,poa/accent text] at (\xx,3.37) {\PoAIcon[2.6mm]{file-text}};
  }
  \node[poa/badge,poa/accent surface,anchor=east,inner xsep=3pt,inner ysep=.6pt] at (16.85,3.37) {\PoAAbuseLowCost};
  \draw[poa/accent arrow] (14.325,3.80) -- (14.325,3.58);
  \draw[poa/accent arrow] (14.325,3.16) -- (14.325,2.94);
  \path[poa/card,draw=PoAAccentBorder] (11.76,2.36) rectangle (17.14,2.94);
  \path[poa/card,draw=PoAAccentBorder] (11.68,2.32) rectangle (17.06,2.90);
  \path[poa/card,poa/accent soft] (11.60,2.28) rectangle (16.98,2.86);
  \node[poa/label,poa/accent text] at (11.80,2.57) {\PoAIcon[2.6mm]{trash}};
  \node[poa/label,poa/accent text,anchor=west] (junk-title) at (12.00,2.57) {\PoAAbuseJunk};
  \node[poa/comparison example,anchor=west] (junk-duplicates)
    at ([xshift=1.5mm]junk-title.east) {\PoAIcon[2.1mm]{repeat-2}\hspace{1.5pt}\PoAAbuseDuplicates};
  \node[poa/comparison example,anchor=west] (junk-spam)
    at ([xshift=1.2mm]junk-duplicates.east) {\PoAIcon[2.1mm]{messages-square}\hspace{1.5pt}\PoAAbuseSpam};

  \draw[poa/accent arrow] (17.14,2.79) -- (17.26,2.79)
    -- (17.26,2.04) -- (15.40,2.04) -- (15.40,1.86);
  \draw[poa/accent arrow] (17.06,2.62) -- (17.15,2.62)
    -- (17.15,2.12) -- (15.20,2.12) -- (15.20,1.86);
  \draw[poa/accent arrow] (16.98,2.45) -- (17.04,2.45)
    -- (17.04,2.20) -- (15.00,2.20) -- (15.00,1.86);

  \path[poa/card,draw=PoARing] (5.30,1.22) rectangle (17.28,1.86);
  \begin{scope}
    \clip[rounded corners=\PoAFigureRadius] (5.30,1.22) rectangle (17.28,1.86);
    \fill[PoAMuted] (5.30,1.22) rectangle (8.98,1.86);
  \end{scope}
  \node[poa/label,anchor=west] at (5.48,1.54) {%
    {\PoAFigureTitleFont\PoAIcon[2.8mm]{server}\hspace{3pt}\PoAModelService}\\
    {\PoAFigureSmallFont\color{PoAFg}\PoAModelServiceExample}%
  };
  \foreach \xx in {8.98,11.7625,14.4975} {
    \draw[draw=PoABorder,line width=\PoAFigureLineWidth] (\xx,1.34) -- (\xx,1.74);}
  \node[poa/label] at (10.395,1.54)
    {\PoAIcon[2.5mm]{list-ordered}\hspace{3pt}\PoAServiceQueue};
  \node[poa/label] at (13.13,1.54)
    {\PoAIcon[2.5mm]{database}\hspace{3pt}\PoAServiceStorage};
  \node[poa/label] at (15.865,1.54)
    {\PoAIcon[2.5mm]{cpu}\hspace{3pt}\PoAServiceCompute};

  \path[poa/card,poa/accent soft] (5.30,.03) rectangle (8.53,1.03);
  \foreach \xx/\dx in {5.23/.28,8.60/-.28} {
    \draw[poa/comparison focus,draw=PoAAccent]
      (\xx,.92) -- (\xx,1.10) -- (\xx+\dx,1.10);
    \draw[poa/comparison focus,draw=PoAAccent]
      (\xx,.14) -- (\xx,-.04) -- (\xx+\dx,-.04);}
  \node[poa/title,poa/accent text,anchor=west] at (5.48,.83)
    {\PoAIcon[2.8mm]{circle-help}\hspace{3pt}\PoAModelGate};
  \node[poa/small,text=PoAFg,anchor=north west,text width=2.87cm] at (5.48,.63) {\PoAModelEvidence};
  \path[poa/card,draw=PoAAccentBorder] (8.70,.03) rectangle (13.77,1.03);
  \node[poa/label,poa/accent text,font=\PoAFigureFont\bfseries,anchor=west] at (8.91,.83) {\PoAModelStatus};
  \node[poa/small,anchor=north west,text width=4.64cm] at (8.91,.63) {\PoAModelQuestion};
  \path[poa/card] (13.94,.03) rectangle (17.28,1.03);
  \node[poa/label,font=\PoAFigureFont\bfseries,anchor=west] at (14.15,.83) {\PoAModelMotivation};
  \node[poa/small,anchor=north west,text width=2.96cm] at (14.15,.63) {\PoAModelProtect};
\end{tikzpicture}%

%% file: sections/02_background.tex
\section{Background and Related Work}
\label{sec:background}

\subsection{Bot Detection and Admission Control}
Bot detection is a mechanism used by Web services to protect against large-scale automated abuse and attacks.
Such systems examine request headers, browser characteristics, and behavior within sessions for signs of automation~\cite{park2006robot,cloudflare2026detection}.
Cloudflare, for example, combines known fingerprints, JavaScript checks, and machine-learning models to assign a bot score indicating how likely a request is to be automated~\cite{cloudflare2026botscore}.
A service may also use an explicit challenge to block automated traffic.
CAPTCHAs follow this approach with tasks designed to be easy for humans and difficult for contemporary programs~\cite{vonahn2003captcha}.
The service checks the submitted response for admission control, admitting clients judged to be human.

Admission control can also make access conditional on computational effort.
Proof-of-work (PoW) raises the cost of large-scale abuse, such as spam, by requiring clients to solve computational problems that are costly to solve but inexpensive to verify~\cite{dwork1993pricing}.
Client puzzles apply this asymmetry to defend against connection-depletion attacks: the server issues a puzzle and allocates connection resources only after verifying a valid solution~\cite{juels1999clientpuzzles}.
Requiring a solution before allocating resources raises the cost of repeated requests, helping protect service capacity from large-scale abuse.

\subsection{LLM-Powered Web Agents}
LLM-powered Web agents are becoming increasingly common on the Internet~\cite{cloudflare2026agenticweb}.
In an LLM-powered Web agent, a language model interprets user instructions and selects actions based on observations from the environment~\cite{gur2024webagent,he2024webvoyager}.
Chain-of-thought (CoT) prompting guides models to reason step by step~\cite{wei2022cot}, while ReAct combines reasoning, actions, and observations~\cite{yao2023react}.
WebAgent brings reasoning to browser interaction by combining planning, HTML understanding, and program generation~\cite{gur2024webagent}.
Moreover, multimodal agents such as WebVoyager and SeeAct incorporate visual observations to interpret webpages and select actions~\cite{he2024webvoyager,zheng2024seeact}.

A Web agent can access Web services through tools such as \texttt{web\_search} and \texttt{web\_fetch}~\cite{anthropicwebsearch,anthropicwebfetch}, MCP interfaces~\cite{mcp2026tools}, browser extensions~\cite{anthropicchrome}, or browser-control and computer-use frameworks~\cite{browseruseactor,anthropiccomputeruse}.
For visual interaction, multimodal agents can observe screenshots of the browser or desktop environment and select actions based on visual cues~\cite{he2024webvoyager,anthropiccomputeruse}.
To evaluate agents' performance on Web tasks, researchers have developed a variety of Web-agent benchmarks.
These include Mind2Web, WebArena, and WorkArena for Web tasks~\cite{deng2023mind2web,zhou2024webarena,drouin2024workarena}, VisualWebArena for visually grounded Web tasks~\cite{koh2024visualwebarena}, and OSWorld for broader computer-use tasks involving Web applications~\cite{xie2024osworld}, showcasing the capabilities of LLM-powered agents across various domains and tasks.

\subsection{Agent Identification and Authentication}
Web platforms are introducing mechanisms to identify agents and authenticate their claimed identities.
Cloudflare, for example, recognizes registered bots and agents through its Verified bots program, while Signed agents supports cryptographically authenticated agent traffic~\cite{cloudflare2026verifiedbots,cloudflare2025signedagents}.
Web Bot Auth checks HTTP request signatures against registered public keys, allowing services to authenticate the sender associated with a claimed agent identity~\cite{cloudflare2026webbotauth}.

Academic research also explores identity and authentication mechanisms for AI agents.
For example, Garzon et al. propose a framework for agents to authenticate one another across organizational boundaries~\cite{garzon2026agentidentity}.
AgentDID uses decentralized identifiers and verifiable credentials to authenticate agent identities~\cite{xu2026agentdid}.
Beyond identity authentication, several studies explore the use of fingerprints to identify AI browsing agents~\cite{wang2026fpagent,fayolle2026llmbot,lugoloobi2026knownactions}, reflecting research interest in this emerging area.
However, identifying an agent or authenticating its identity does not by itself establish whether a language model participates in the current interaction.
On the human-facing Web, account authentication verifies control of an account, but valid credentials can also be used by automated programs.
CAPTCHAs and other bot detection mechanisms therefore seek evidence of human participation to limit automated abuse~\cite{vonahn2003captcha,cloudflare2026botscore}.
Similarly, an authenticated agent identity can also be used by a scripted program.
This motivates a complementary mechanism for verifying model participation in agent-facing Web services.

%% file: sections/03_problem.tex
\section{Proof-of-Agentness: Definition and Design}
\label{sec:problem}

\subsection{Problem Formulation}
Drawing inspiration from CAPTCHAs and bot detection, we propose \textit{Proof-of-Agentness} (\PoA{}) based on the resource analogy between human attention and language model inference.
Human participation consumes time and attention, while language model participation consumes inference compute and tokens.
Simple scripts can repeatedly submit requests at low marginal cost while occupying server-side queues and consuming storage and computing resources.
\PoA{} is designed to extend the principle of admitting humans while keeping bots out to agent-facing Web services.
The goal is to admit agents while keeping automated bots out, empowering Web services to regulate automated traffic more selectively.

\begin{poadefinition}[title={Definition: Proof-of-Agentness}]
\textit{Proof-of-Agentness} (\PoA{}) is a mechanism for verifying \textit{language model participation} in an interaction.
Participation means that model output is used to interpret information, select an action, or produce a response in that interaction.
\end{poadefinition}

Requiring evidence of model participation could help agent-facing services prevent automated bots from gaining access by posing as agents, limiting opportunities for large-scale automated abuse at no cost.
Such a mechanism can be deployed across Web interfaces, APIs, and MCP tools, particularly for operations that change service state, allocate limited resources, or directly affect other users.
Services can specify which interactions or operations require evidence of model participation, making participation an explicit part of their access policies.

\subsection{Design Objectives}
Beyond verifying agent participation, \PoA{} should also remain practical for services and agents.
This requires considering both service overhead and agent effort during ordinary interactions.
We identify three practical design goals for deployment.

\begin{itemize}
  \item \textbf{Efficiency.}
  Verification should impose limited computation, storage, and communication costs on the service, allowing it to scale with incoming traffic.
  \item \textbf{Validity.}
  Verification results should be bound to the client and permitted operations, with explicit expiry and reuse limits.
  The service should enforce these limits so that results cannot authorize access outside their intended scope.
  \item \textbf{Usability.}
  Verification requirements and formats should be compatible with supported agents and their access interfaces.
  The additional latency, inference cost, and interaction effort should remain acceptable for useful agent work.
\end{itemize}

In pursuing these objectives, \PoA{} faces a familiar constraint: a service generally observes a client's interactions without direct access to the computation behind them.
A similar challenge is faced by CAPTCHAs and bot detection mechanisms, whose design principles and deployment experience offer a foundation for extending participation verification to language models.

\subsection{Verification Framework}
\label{sec:verification-framework}
\label{sec:binding}
\begin{figure*}[t]
  \centering
  \input{figures/02_framework.tex}
  \caption{Agent verification framework.
  The proposed framework combines active and passive evidence with verified agent declarations through a unified \AgentScore{} to guide access to protected operations.}
  \Description{Active verification and passive analysis feed a unified AgentScore through policy rules and fitted passive scores.}
  \label{fig:evidence-routes}
\end{figure*}
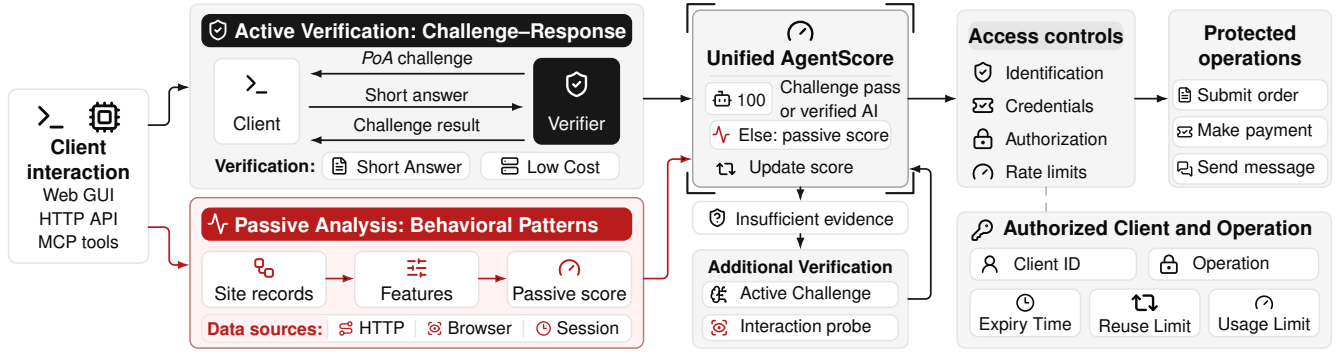

Building on these principles and prior work, we organize \PoA{} around two approaches to obtaining participation evidence: active verification and passive analysis.
Figure~\ref{fig:evidence-routes} presents our framework, linking evidence to access decisions.

\noindent\textbf{Active verification.}
The service issues a public challenge $c$ and retains private verification state $\sigma$.
After receiving the client's answer $\widehat a$, it computes $a=\operatorname{Verify}(\sigma,\widehat a)$, where $a\in\{0,1\}$ indicates failure or success.
The challenge is designed under the assumption that it is easy for language models to solve but difficult for simple rule-based programs.
Under this assumption, a correct response provides evidence of agent participation, and the strength of this evidence depends on the soundness of the challenge design.

\noindent\textbf{Passive analysis.}
During ordinary use, the service collects a record $h$ of observable interactions.
A feature extractor maps the available observations to $x=\Phi(h)$, capturing patterns in requests and browser interactions.
This approach relies on the assumption that agents can produce observable behavioral differences from other automated clients.
A model trained on labeled interactions uses these features to estimate \(p_A(x)\), the probability that the client is an Agent.
The strength of this evidence depends on the quality of observed signals and the reliability of behavioral analysis.

\noindent\textbf{Hybrid policy.}
In practice, a service could use a hybrid policy to combine multiple sources of evidence, such as active challenge results, passive interaction signals, and verified agent declarations~\cite{cloudflare2026webbotauth}.
The policy could also incorporate existing access controls to govern access to protected operations.
To support such hybrid policies, we introduce \AgentScore{}, a unified score expressing the service's confidence in recognizing a client as an Agent under its policy.
If the score is too low to support the requested access, the service may defer its decision and request additional verification through an active challenge or an \textit{interaction probe}.
The resulting evidence updates \AgentScore{}, allowing the service to reassess the request.
Moreover, the permission granted should be bound to the authorized client and permitted operations, with expiry and usage limits.
We detail the proposed scoring and access policy in Section~\ref{sec:agent-score}.

%% file: figures/02_framework.tex
\newcommand{\PoAEvidenceInput}{Client\\interaction}%
\newcommand{\PoAEvidenceInputBody}{Web GUI\\HTTP API\\MCP tools}%
\newcommand{\PoAEvidenceRule}{Active Verification: Challenge--Response}%
\newcommand{\PoAEvidencePassive}{Passive Analysis: Behavioral Patterns}%
\newcommand{\PoAEvidenceFusion}{Unified AgentScore}%
\newcommand{\PoAEvidenceOutput}{Access controls}%
\newcommand{\PoAEIdentification}{Identification}%
\newcommand{\PoAECredentials}{Credentials}%
\newcommand{\PoAEAuthorization}{Authorization}%
\newcommand{\PoAERateLimits}{Rate limits}%
\newcommand{\PoAEVerifier}{Verifier}%
\newcommand{\PoAEClient}{Client}%
\newcommand{\PoAEQuestion}{\PoA{} challenge}%
\newcommand{\PoAEAnswer}{Short answer}%
\newcommand{\PoAEResult}{Challenge result}%
\newcommand{\PoAEVerification}{Verification:}%
\newcommand{\PoAEShortAnswer}{Short Answer}%
\newcommand{\PoAELowCost}{Low Cost}%
\newcommand{\PoAERecords}{Site records}%
\newcommand{\PoAEFeatures}{Features}%
\newcommand{\PoAELearner}{Passive score}%
\newcommand{\PoAEDataSources}{Data sources:}%
\newcommand{\PoAEHTTPRecords}{HTTP}%
\newcommand{\PoAEBrowserRecords}{Browser}%
\newcommand{\PoAESessionRecords}{Session}%
\newcommand{\PoAERuleMatched}{Challenge pass\\or verified AI}%
\newcommand{\PoAERuleValue}{100}%
\newcommand{\PoAEPassiveFallback}{Else: passive score}%
\newcommand{\PoAEUpdateScore}{Update score}%
\newcommand{\PoAEOperation}{Protected\\operations}%
\newcommand{\PoAEOrder}{Submit order}%
\newcommand{\PoAEPayment}{Make payment}%
\newcommand{\PoAEMessage}{Send message}%
\newcommand{\PoAEScope}{Authorized Client and Operation}%
\newcommand{\PoAEClientIdentity}{Client ID}%
\newcommand{\PoAEAllowedOperation}{Operation}%
\newcommand{\PoAEExpiry}{Expiry Time}%
\newcommand{\PoAEReuse}{Reuse Limit}%
\newcommand{\PoAEUsage}{Usage Limit}%
\newcommand{\PoAEInsufficient}{Insufficient evidence}%
\newcommand{\PoAEAdditional}{Additional Verification}%
\newcommand{\PoAEActiveChallenge}{Active Challenge}%
\newcommand{\PoAEInteractionProbe}{Interaction probe}%

\begin{tikzpicture}[poa/picture]
  \path[use as bounding box] (0,0) rectangle (17.5,4.6);
  \tikzset{
    poa/evidence feature/.style={poa/card, font=\PoAFigureSmallFont,
      minimum height=.36cm, inner sep=0pt},
    poa/evidence source/.style={poa/label, font=\PoAFigureSmallFont},
    poa/evidence route title/.style={poa/title,poa/inverse text,
      anchor=west,outer sep=0pt}
  }

  \path[poa/card] (0,1.175) rectangle (1.85,3.475);
  \node at (.56,3.105) {\PoAIcon[4.5mm]{terminal}};
  \node at (1.27,3.105) {\PoAIcon[4.5mm]{cpu}};
  \node[poa/title,align=center] at (.925,2.545) {\PoAEvidenceInput};
  \node[poa/small,align=center] at (.925,1.765) {\PoAEvidenceInputBody};
  \draw[poa/arrow] (1.85,3.00) -- (2.125,3.00) |- (2.4,3.50);
  \draw[poa/accent arrow] (1.85,1.65) -- (2.125,1.65) |- (2.4,1.15);

  \path[poa/panel] (2.4,2.16) rectangle (8.4,4.6);
  \path[poa/card,poa/dark surface] (2.56,4.04) rectangle (8.24,4.46);
  \node[poa/evidence route title] (active-route-title) at (2.62,4.25)
    {\PoAIcon[3mm]{shield-check}\hspace{2pt}\PoAEvidenceRule};
  \path[poa/card] (2.72,2.75) rectangle (3.85,3.88);
  \node at (3.285,3.54) {\PoAIcon[3.7mm]{terminal}};
  \node[poa/label] at (3.285,3.02) {\PoAEClient};
  \path[poa/card,poa/dark surface] (6.95,2.75) rectangle (8.08,3.88);
  \node[poa/inverse text] at (7.515,3.54) {\PoAIcon[3.7mm]{shield-check}};
  \node[poa/label,poa/inverse text] at (7.515,3.02) {\PoAEVerifier};
  \foreach \yy/\poaEvidenceLabel in {3.68/PoAEQuestion,2.80/PoAEResult} {
    \draw[poa/arrow] (6.82,\yy) -- (3.98,\yy);
    \node[poa/small,anchor=south] at (5.4,\yy+.07) {\csname\poaEvidenceLabel\endcsname};}
  \draw[poa/arrow] (3.98,3.24) -- (6.82,3.24);
  \node[poa/small,anchor=south] at (5.4,3.31) {\PoAEAnswer};
  \node[poa/small,text=PoAFg,font=\PoAFigureSmallFont\bfseries,anchor=west]
    at (2.72,2.46) {\PoAEVerification};
  \node[poa/evidence feature,minimum width=1.95cm] at (5.125,2.46)
    {\PoAIcon[2.5mm]{file-text}\hspace{3pt}\PoAEShortAnswer};
  \node[poa/evidence feature,minimum width=1.83cm] at (7.165,2.46)
    {\PoAIcon[2.5mm]{server}\hspace{3pt}\PoAELowCost};

  \path[poa/panel,poa/accent soft] (2.4,.05) rectangle (8.4,2.04);
  \path[poa/card,poa/accent surface] (2.56,1.48) rectangle (8.24,1.90);
  \node[poa/evidence route title] (passive-route-title) at (2.62,1.69)
    {\PoAIcon[3mm]{activity}\hspace{2pt}\PoAEvidencePassive};
  \foreach \xx/\ic/\poaEvidenceLabel in {3.38/workflow/PoAERecords,5.4/sliders-horizontal/PoAEFeatures,7.42/gauge/PoAELearner} {
    \path[poa/card] (\xx-.82,.59) rectangle (\xx+.82,1.33);
    \node[poa/accent text] at (\xx,1.11) {\PoAIcon[3.1mm]{\ic}};
    \node[poa/label] at (\xx,.77) {\csname\poaEvidenceLabel\endcsname};}
  \draw[poa/accent arrow] (4.20,.97) -- (4.58,.97);
  \draw[poa/accent arrow] (6.22,.97) -- (6.60,.97);
  \path[poa/card] (2.56,.13) rectangle (8.24,.49);
  \begin{scope}
    \clip[rounded corners=\PoAFigureRadius] (2.56,.13) rectangle (8.24,.49);
    \fill[PoAAccentSoft] (2.56,.13) rectangle (4.22,.49);
  \end{scope}
  \node[poa/small,poa/accent text,font=\PoAFigureSmallFont\bfseries]
    at (3.39,.31) {\PoAEDataSources};
  \foreach \xx in {4.22,5.38,6.81} {
    \draw[draw=PoABorder,line width=\PoAFigureLineWidth] (\xx,.20) -- (\xx,.42);}
  \node[poa/evidence source] at (4.80,.31)
    {{\color{PoAAccent}\PoAIcon[2.1mm]{route}}\hspace{2pt}\PoAEHTTPRecords};
  \node[poa/evidence source] at (6.095,.31)
    {{\color{PoAAccent}\PoAIcon[2.1mm]{scan-eye}}\hspace{2pt}\PoAEBrowserRecords};
  \node[poa/evidence source] at (7.525,.31)
    {{\color{PoAAccent}\PoAIcon[2.1mm]{clock-3}}\hspace{2pt}\PoAESessionRecords};

  \draw[poa/arrow] (8.4,3.35) -- (9.05,3.35);
  \draw[poa/accent arrow] (8.4,.97) -- (8.72,.97) |- (9.05,2.55);
  \path[poa/panel,draw=PoARing] (9.05,2.18) rectangle (11.9,4.52);
  \foreach \xx/\dx in {8.98/.36,11.97/-.36} {
    \draw[draw=PoAPrimary,line width=.8pt,rounded corners=0pt]
      (\xx,4.23) -- (\xx,4.60) -- (\xx+\dx,4.60);
    \draw[draw=PoAPrimary,line width=.8pt,rounded corners=0pt]
      (\xx,2.37) -- (\xx,2.10) -- (\xx+\dx,2.10);}
  \node at (10.475,4.24) {\PoAIcon[3.8mm]{gauge}};
  \node[poa/title] at (10.475,3.86) {\PoAEvidenceFusion};
  \draw[draw=PoABorder,line width=.55pt] (9.28,3.65) -- (11.67,3.65);
  \node[poa/badge,fill=PoABackground,minimum width=.88cm,inner xsep=2pt] at (9.66,3.35)
    {\PoAIcon[2.8mm]{bot}\hspace{1.5pt}\PoAERuleValue};
  \node[poa/small,text=PoAFg,anchor=west] at (10.20,3.35) {\PoAERuleMatched};
  \path[poa/card] (9.28,2.68) rectangle (11.67,3.06);
  \node[poa/small,text=PoAFg] at (10.475,2.87) {{\color{PoAAccent}\PoAIcon[2.5mm]{activity}}\hspace{3pt}\PoAEPassiveFallback};
  \node at (9.49,2.42) {\PoAIcon[2.8mm]{repeat-2}};
  \node[poa/small,text=PoAFg,anchor=west] at (9.79,2.42) {\PoAEUpdateScore};
  \draw[poa/arrow] (10.475,2.18) -- (10.475,1.99);
  \path[poa/card] (9.05,1.56) rectangle (11.9,1.99);
  \node[poa/small,text=PoAFg] at (10.475,1.775) {\PoAIcon[2.7mm]{shield-question-mark}\hspace{3pt}\PoAEInsufficient};

  \draw[poa/arrow] (10.475,1.56) -- (10.475,1.35);
  \path[poa/panel] (9.05,.05) rectangle (11.9,1.35);
  \node[poa/small,text=PoAFg,font=\PoAFigureSmallFont\bfseries] at (10.475,1.13) {\PoAEAdditional};
  \path[poa/card] (9.18,.59) rectangle (11.77,.93);
  \node at (9.40,.76) {\PoAIcon[2.5mm]{brain-circuit}};
  \node[poa/small,text=PoAFg,anchor=west] at (9.67,.76) {\PoAEActiveChallenge};
  \path[poa/card] (9.18,.15) rectangle (11.77,.49);
  \node[poa/accent text] at (9.40,.32) {\PoAIcon[2.5mm]{scan-eye}};
  \node[poa/small,text=PoAFg,anchor=west] at (9.67,.32) {\PoAEInteractionProbe};
  \draw[poa/arrow] (11.9,.70) -- (12.20,.70) -- (12.20,2.42) -- (11.9,2.42);

  \draw[poa/arrow] (11.9,3.35) -- (12.55,3.35);
  \path[poa/panel] (12.55,2.18) rectangle (14.9,4.52);
  \path[poa/card,fill=PoABorder] (12.71,4.00) rectangle (14.74,4.36);
  \node[poa/title] at (13.725,4.18) {\PoAEvidenceOutput};
  \foreach \yy/\ic/\poaEvidenceLabel in {3.70/shield-check/PoAEIdentification,3.26/ticket-check/PoAECredentials,2.82/lock-keyhole/PoAEAuthorization,2.38/gauge/PoAERateLimits} {
    \node at (12.89,\yy) {\PoAIcon[3mm]{\ic}};
    \node[poa/small,text=PoAFg,anchor=west] at (13.18,\yy) {\csname\poaEvidenceLabel\endcsname};}
  \draw[poa/arrow] (14.9,3.35) -- (15.35,3.35);
  \path[poa/panel] (15.35,2.18) rectangle (17.5,4.52);
  \node[poa/title,align=center] at (16.425,4.03) {\PoAEOperation};
  \foreach \yy/\ic/\poaEvidenceLabel in {3.40/file-text/PoAEOrder,2.91/ticket-check/PoAEPayment,2.42/messages-square/PoAEMessage} {
    \node[poa/card,minimum width=2.05cm,minimum height=.40cm,inner sep=0pt] at (16.425,\yy) {};
    \node[inner sep=0pt] at (15.56,\yy) {\PoAIcon[2.2mm]{\ic}};
    \node[poa/small,text=PoAFg,anchor=west,text width=1.70cm] at (15.72,\yy) {\csname\poaEvidenceLabel\endcsname};}

  \path[poa/panel] (12.55,.05) rectangle (17.5,1.85);
  \node at (12.88,1.61) {\PoAIcon[3.1mm]{key-round}};
  \node[poa/title,anchor=west] at (13.15,1.61) {\PoAEScope};
  \path[poa/card] (12.72,.95) rectangle (14.92,1.37);
  \node at (12.98,1.16) {\PoAIcon[2.8mm]{user-round}};
  \node[poa/small,text=PoAFg,anchor=west] at (13.29,1.16) {\PoAEClientIdentity};
  \path[poa/card] (15.08,.95) rectangle (17.33,1.37);
  \node at (15.35,1.16) {\PoAIcon[2.8mm]{lock-keyhole}};
  \node[poa/small,text=PoAFg,anchor=west] at (15.66,1.16) {\PoAEAllowedOperation};
  \foreach \xx/\ic/\poaEvidenceIconSize/\poaEvidenceLabel in {13.45/clock-3/2.6mm/PoAEExpiry,15.03/repeat-2/3.8mm/PoAEReuse,16.61/gauge/2.6mm/PoAEUsage} {
    \path[poa/card] (\xx-.73,.20) rectangle (\xx+.73,.83);
    \node at (\xx,.65) {\PoAIcon[\poaEvidenceIconSize]{\ic}};
    \node[poa/small,text=PoAFg] at (\xx,.35) {\csname\poaEvidenceLabel\endcsname};}
  \draw[draw=PoARing,line width=.5pt,dashed] (13.725,2.18) -- (13.725,1.85);
\end{tikzpicture}%

%% file: sections/04_design.tex
\section{Implementation}
\label{sec:design}
In this section, we present our implementation of the active and passive approaches and detail the proposed \AgentScore{} policy.

\subsection{Active Verification}
\label{sec:challenge}
\label{sec:document-tasks}
We implement active verification through challenge--response based on \textit{natural language understanding} (NLU), a basic capability of language models.
NLU tasks assess the interpretation of text and its implications~\cite{wang2019glue,nie2020adversarial}, and carefully designed questions can expose the limitations of shallow rules and pattern matching~\cite{mccoy2019right,levesque2012winograd}.
Structured simulations and templates also support the systematic generation of natural-language questions and test cases~\cite{weston2016babi,ribeiro2020checklist}.

Building on these studies, we consider NLU a plausible basis for challenges intended to be easy for agents yet difficult for bots.
We also believe that other forms of challenge such as mathematical reasoning and visual understanding have potential for agent verification.
Nonetheless, we use NLU as a starting point and develop a methodology for dynamically constructing such challenges.

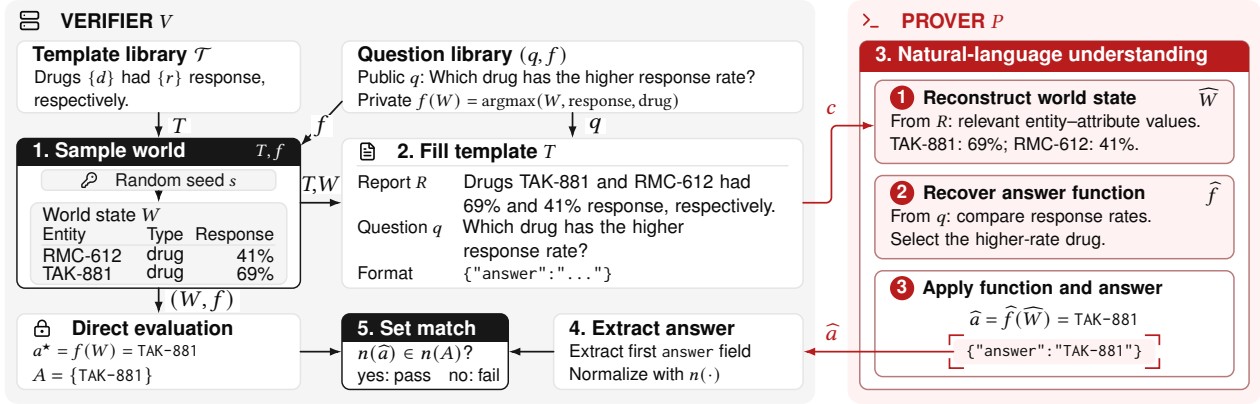
\begin{figure*}[t]
  \centering
  \input{figures/03_challenge.tex}%
  \caption{Structured generation and semantic challenge solving.
  Prepared templates and question pairs $(q,f)$ let the verifier render a report from sampled state $W$ and compute $a^\star=f(W)$ directly.
  The prover uses the public challenge $c=(R,q)$ to reconstruct relevant world-state facts and infer the answer function's semantics, then returns an answer for verification.}
  \Description{The verifier prepares template and question libraries, samples a world state from a random seed, fills a narrative template, and directly evaluates a retained query to obtain reference answers.
  The prover receives the public report and question, reconstructs relevant world-state facts and recovers the answer function's semantics, then returns an answer for extraction and normalized set matching; the structured state and executable query remain with the verifier.}
  \label{fig:active-challenge}
\end{figure*}

This methodology begins with crafting narrative templates and their accompanying questions.
Each template is a long passage with slots for entity names, attribute names, and values, and the questions contain corresponding slots.
Each question has a natural-language form for presentation and a corresponding executable function $f$ for computing its answer once the slots are instantiated.
For each challenge, a \textit{verifier} uses a random seed to sample entities and attribute values from predefined sets, forming a \textit{world state} $W$ that records the sampled entities and values.
It uses this state to fill a selected narrative template and its question, producing a natural-language report $R$ and question $q$.
The verifier applies the function $f$ associated with $q$ to $W$ to compute the answer directly.
The responding \textit{prover} receives only the natural-language report and question.
It is expected to reconstruct the relevant facts from the report and infer the question's intended operation to derive its answer.
Figure~\ref{fig:active-challenge} illustrates this construction.
We describe the protocol between verifier $V$ and prover $P$ as a tuple of four procedures, $\Pi=(\text{\textsc{Setup}},\text{\textsc{Gen}},\text{\textsc{Interact}},\text{\textsc{Verify}})$:

\begin{enumerate}[label=(\arabic*)]
  \item \textsc{Setup}$()\to\mathcal L$.
  First, prepare a resource library $\mathcal L$ containing narrative templates, paired questions, typed entities, and permitted attribute values.
  The verifier should keep this library $\mathcal L$ private because knowledge of its contents could allow bots to solve the challenges using fixed rules.

  \item \textsc{Gen}$(\mathcal L;s)\to(c,\sigma)$.
  Then, using a fresh private seed $s$, the verifier selects a template and samples entities and attribute values to form a \textit{world state} $W$.
  It fills the template and its associated question with these facts, producing report $R$ and question $q$.
  The function $f$ associated with question $q$ computes the answer $a^\star=f(W)$, which the verifier uses to construct an \textit{accepted-answer set} $A$ for verification.
  The procedure outputs the \textit{public challenge} $c=(R,q)$ and retains private verification state $\sigma=(s,W,f,A)$.

  \item \textsc{Interact}$(c)\leftrightarrow P$.
  The prover receives $c=(R,q)$ and returns its answer $\widehat a$ in the required format.
  It is expected to recover the relevant facts and intended operation by interpreting the report and question.
  \item \textsc{Verify}$(\sigma,\widehat a)\to\{0,1\}$.
  The verifier extracts and normalizes the submitted answer and compares it against the retained set $A$, returning 1 on a match and 0 otherwise.
\end{enumerate}

\noindent\textbf{Security assumptions.}
The construction relies on the assumption that language models can solve the generated challenges reliably, while simple rule-based programs cannot achieve comparable success.
If this assumption holds, a correct response provides evidence of agent participation.
Security also depends on keeping each random seed $s$ and the resource library $\mathcal L$ private.
Knowledge of both could allow an adversary to reproduce the \textit{world state} and compute the answer directly.
Repeated observations of public challenges may also reveal the library's templates and question functions.
The verifier should therefore periodically update or replace $\mathcal L$ to make reconstruction from accumulated observations more difficult.

We present the evaluation results of this approach in Section~\ref{sec:challenge-evaluation} and provide further implementation details and a worked challenge example in Appendix~\ref{app:challenge-details}.

\subsection{Passive Analysis}
\label{sec:behavior}
Existing bot detection studies characterize human and bot traffic through request characteristics, browser fingerprints, and interaction behavior~\cite{park2006robot,jonker2019fingerprint,iliou2021advancedbots}.
A prominent industrial deployment is Cloudflare Bot Management, which combines several detection engines to assign bot scores~\cite{cloudflare2026botscore}.
Its Precursor system further analyzes interaction signals throughout a session~\cite{cloudflare2026precursor}.

\begin{table}[t]
  \PoAPreviewTableStyle
  \setlength{\tabcolsep}{\PoAAccessTableColumnSep}
  \begin{threeparttable}
    \caption{Web access across humans, bots, and agents.
      The table summarizes representative interfaces and tools typically used by humans, bots, and LLM-powered agents, their possible operations, and the signals observable by Web services.}
    \label{tab:web-access-signals}
    \input{tables/01_access.tex}
    \begin{tablenotes}[normal]
      \footnotesize
      \raggedright
      \item[a] \textbf{Typical user types:} \PoAIcon[\PoATableRoleIconSize]{user-round} Human; \textcolor{PoAMutedFg}{\PoAIcon[\PoATableRoleIconSize]{bot}} Bot; \textcolor{PoAAccent}{\PoAIcon[\PoATableRoleIconSize]{brain-circuit}} Agent.
      Icons indicate typical users; 
      users can use other interfaces too; for example, humans can also use curl.
      \item[b] \textbf{Possible operations:} R: read; W: write or change service state.
      Final available operations depend on the interface and permissions.
      \item[c] \textbf{Observable signals:} H: HTTP; B: browser environment; E: page events.
      B/E require page execution and data collection from the browser.
      \item[d] \textbf{Var.:} Observable signals depend on the underlying tool implementation.
    \end{tablenotes}
  \end{threeparttable}
\end{table}

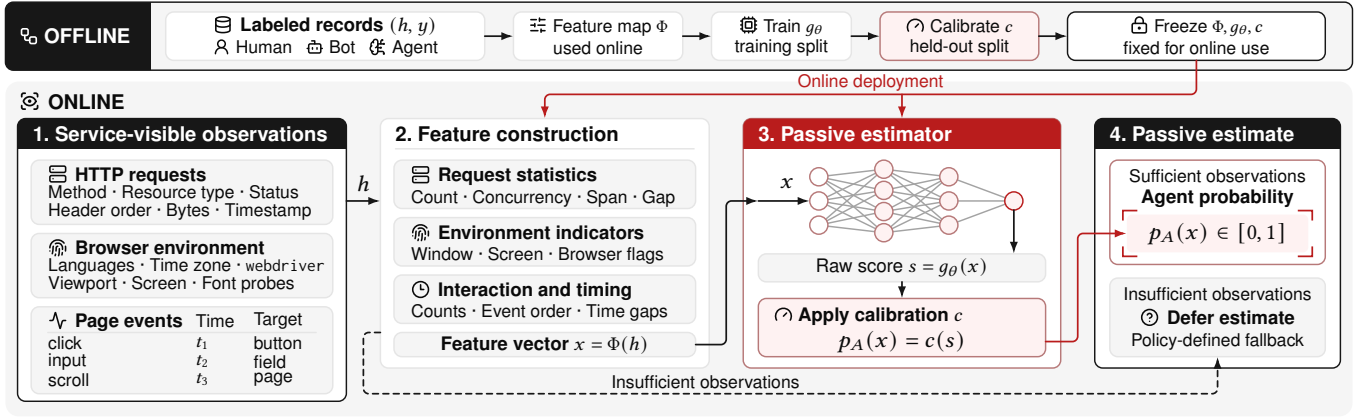
\begin{figure*}[t]
  \centering
  \input{figures/04_passive.tex}
  \caption{Passive analysis pipeline.
  Offline training and proposed calibration prepare a fixed predictor for online Agent-probability estimation; insufficient observations defer estimation.}
  \Description{Offline processing uses labeled Human, Bot, and Agent records to define shared features, train a classifier, and fit a calibration function on held-out data before deployment.
  Online HTTP, browser, and page-event observations yield an Agent probability; insufficient observations defer estimation, with subsequent handling determined by service policy.}
  \label{fig:passive-analysis}
\end{figure*}

However, as LLM-powered agents become participants in Web traffic, it is still unclear whether they exhibit observable patterns distinct from those of humans and traditional bots.
We started by reviewing Web access approaches available to agents in terms of their supported operations and service-visible signals, as summarized in Table~\ref{tab:web-access-signals}.
Agents can use nearly all the access approaches listed, sharing interfaces with both humans and traditional bots.
Their observable patterns can vary across interfaces and execution tools, overlapping with those of other client types.
Automated scripts may also use agent-oriented interfaces and frameworks, such as MCP and Browser Use~\cite{mcp2026tools,browseruseagents}, to pose as agents.
Agent verification on the Web fundamentally involves making a finer distinction between agents and bots within automated traffic.
Reliable passive analysis generally relies on large, diverse, and representative datasets, which may give providers with broad visibility into Web traffic an advantage in developing and maintaining such models.
We illustrate the passive analysis pipeline in Figure~\ref{fig:passive-analysis}.

Here, we created a minimal dataset containing Web interactions from human participants, traditional bots, and LLM-powered agents, together with associated client fingerprints.
Then, we conduct a preliminary evaluation using supervised classification methods to explore the feasibility of this approach.
We present the experimental details and results in Section~\ref{sec:evaluation}.

\subsection{Hybrid Scoring Policy}
\label{sec:agent-score}
We detail how a service could integrate the available evidence into an \AgentScore{}.
Similar to Cloudflare's Bot Score~\cite{cloudflare2026botscore}, \AgentScore{} denotes the service's confidence in recognizing a client as an agent and can be used alongside existing access controls to govern protected operations.

\noindent\textbf{Unified score.}
We express the available evidence as an \AgentScore{} on a common scale in $[0,100]$.
Let $r\in\{0,1\}$ indicate whether a client passes an active challenge or presents a verified agent declaration~\cite{cloudflare2026webbotauth}, and let $u\in\{0,1\}$ indicate whether sufficient observations for passive analysis are available.
The proposed score is
\begin{equation}
  S(x,r,u)=
  \begin{cases}
    100, & r=1, \\
    100\,p_A(x), & r=0 \text{ and } u=1, \\
    \bot, & \text{otherwise}.
  \end{cases}
  \label{eq:unified-agent-score}
\end{equation}
The \textit{rule-assigned score} of 100 expresses the service's decision to accept the verified evidence under its policy.
For the passive branch, $p_A(x)$ from Section~\ref{sec:behavior} is scaled to $100p_A(x)$.
Here, $\bot$ means that neither verified rule evidence nor a usable passive estimate is available, leaving further handling to the service's policy.

\noindent\textbf{Access decision.}
\AgentScore{} could be used alongside identification and authorization to inform access decisions.
For each protected operation $o$, let $\tau_o\in[0,100]$ be the required score and $C_\pi(i,o)\in\{0,1\}$ indicate whether client $i$ satisfies these controls under policy $\pi$.
For score $s=S(x,r,u)$, the final access decision $d$ can be expressed as
\begin{equation}
  d=\mathrm{allow}\;\Longleftrightarrow\;s\in[\tau_o,100]\land C_\pi(i,o)=1.
  \label{eq:access-decision}
\end{equation}
Otherwise, the policy determines whether to deny or defer the request.
If $s=\bot$ or $s<\tau_o$, the service may request an active challenge or an interaction probe, such as a Turnstile-like embedded webpage component~\cite{cloudflare2026turnstile}, to update \AgentScore{} and reassess access.
Granted permissions should also be bound to the authorized client and permitted operations, with expiry and usage limits.
Together, these mechanisms could give Web services finer control over automated access: allowing AI agents to act on behalf of users while limiting large-scale scripted abuse of service resources.

%% file: figures/03_challenge.tex
\newcommand{\PoASRServerRole}{VERIFIER $V$}%
\newcommand{\PoASRClientRole}{PROVER $P$}%
\newcommand{\PoASRWorldTitle}{1. Sample world}%
\newcommand{\PoASRSeed}{Random seed $s$}%
\newcommand{\PoASRWorldLabel}{World state $W$}%
\newcommand{\PoASRPublicTitle}{2. Fill template $T$}%
\newcommand{\PoASRReport}{Drugs TAK-881 and RMC-612 had\\69\% and 41\% response, respectively.}%
\newcommand{\PoASRQuestion}{Which drug has the higher\\response rate?}%
\newcommand{\PoASRFormat}{\texttt{\{"answer":"..."\}}}%
\newcommand{\PoASRClientTitle}{3. Natural-language understanding}%
\newcommand{\PoASRFactsTitle}{Reconstruct world state}%
\newcommand{\PoASRFactsBody}{From $R$: relevant entity--attribute values.\\
  TAK-881: 69\%; RMC-612: 41\%.}%
\newcommand{\PoASRQueryTitle}{Recover answer function}%
\newcommand{\PoASRQueryBody}{From $q$: compare response rates.\\
  Select the higher-rate drug.}%
\newcommand{\PoASRCompareTitle}{Apply function and answer}%
\newcommand{\PoASRCompareBody}{$\widehat a=\widehat f(\widehat W)=\texttt{TAK-881}$}%
\newcommand{\PoASRClientBody}{\texttt{\{"answer":"TAK-881"\}}}%
\newcommand{\PoASRNormalizeTitle}{4. Extract answer}%
\newcommand{\PoASRNormalizeBody}{Extract first \texttt{answer} field\\Normalize with $n(\cdot)$}%
\newcommand{\PoASRCheckTitle}{5. Set match}%
\newcommand{\PoASRCheckBody}{$n(\widehat a)\in n(A)$?}%
\newcommand{\PoASRResult}{yes: pass\quad no: fail}%
\newcommand{\PoASRReferenceTitle}{Direct evaluation}%
\newcommand{\PoASRReference}{$A=\{\texttt{TAK-881}\}$}%
\newcommand{\PoASRDerive}{$a^\star=f(W)=\texttt{TAK-881}$}%

\begin{tikzpicture}[poa/picture]
  \path[use as bounding box] (-.16,.30) rectangle (16.46,5.64);
  \tikzset{
    poa/sr card/.style={poa/card, inner sep=0pt},
    poa/sr role/.style={poa/title, anchor=west},
    poa/sr title/.style={poa/title, anchor=west},
    poa/sr body/.style={poa/label, anchor=north west},
    poa/sr transfer/.style={poa/label, font=\PoAFigureSymbolFont,
      align=center, fill=PoABackground, inner xsep=1.5pt, inner ysep=0pt},
    poa/sr prover title/.style={poa/label, font=\PoAFigureFont\bfseries,
      anchor=base west, text width=3.55cm},
    poa/sr step/.style={circle, fill=PoAAccent, text=PoAOnDark,
      font=\PoAFigureFont\bfseries, minimum size=3.2mm, inner sep=0pt,
      outer sep=0pt, text height=1.2ex, text depth=.2ex},
    poa/sr prover symbol/.style={poa/label, font=\PoAFigureFont,
      anchor=base east, yshift=-.3mm},
    poa/sr prover body/.style={poa/small,anchor=north west,text width=4.35cm},
    poa/sr focus/.style={draw=PoAAccent, line width=\PoAFigureArrowWidth,
      line cap=rect, line join=miter, rounded corners=0pt}
  }

  \path[fill=PoAMuted,draw=none,rounded corners=5pt]
    (-.16,.30) rectangle (10.56,5.64);
  \path[fill=PoAAccentSoft,draw=none,rounded corners=5pt]
    (10.99,.30) rectangle (16.46,5.64);

  \node[poa/sr role] at (0.02,5.37)
    {\PoAIcon[2.8mm]{server}\quad\PoASRServerRole};
  \node[poa/sr role, text=PoAAccent] at (11.15,5.37)
    {\PoAIcon[2.8mm]{terminal}\quad\PoASRClientRole};

  \draw[poa/sr card] (0,4.15) rectangle (3.75,5.10);
  \node[poa/sr title] at (0.20,4.90) {Template library $\mathcal T$};
  \node[poa/small, anchor=base west, text width=3.33cm] at (0.22,4.55)
    {Drugs $\{d\}$ had $\{r\}$ response,\\respectively.};
  \draw[poa/sr card] (4.30,4.15) rectangle (10.40,5.10);
  \node[poa/sr title] at (4.50,4.90) {Question library $(q,f)$};
  \node[poa/small, anchor=base west] at (4.50,4.55)
    {Public $q$: Which drug has the higher response rate?};
  \node[poa/small, anchor=base west] at (4.50,4.24)
    {Private $f(W)=\operatorname{argmax}(W,\mathrm{response},\mathrm{drug})$};
  \draw[poa/arrow] (1.875,4.15) -- (1.875,3.81);
  \node[poa/sr transfer] at (2.15,3.98) {$T$};
  \draw[poa/arrow] (4.30,4.30) -- (3.75,3.71);
  \node[poa/sr transfer] at (4.03,3.98) {$f$};
  \draw[poa/arrow] (7.35,4.15) -- (7.35,3.81);
  \node[poa/sr transfer] at (7.65,3.98) {$q$};

  \draw[poa/sr card, draw=PoAPrimary] (0,1.83) rectangle (3.75,3.81);
  \begin{scope}
    \clip[rounded corners=\PoAFigureRadius] (0,1.83) rectangle (3.75,3.81);
    \fill[PoAPrimary] (0,3.45) rectangle (3.75,3.81);
  \end{scope}
  \node[poa/sr title, poa/inverse text] at (0.20,3.63) {\PoASRWorldTitle};
  \node[poa/small, poa/inverse text, anchor=east] at (3.55,3.63) {$T,f$};
  \node[poa/badge, minimum width=3.10cm, minimum height=0.27cm,inner ysep=.5pt]
    (sr-world-seed) at (1.875,3.26) {\PoAIcon[2.3mm]{key-round}\quad\PoASRSeed};
  \draw[poa/arrow] (sr-world-seed.south) -- (1.875,2.96);
  \draw[poa/sr card, fill=PoAMuted] (0.18,1.88) rectangle (3.57,2.96);
  \node[poa/label, anchor=west] at (0.33,2.80) {\PoASRWorldLabel};
  \node[poa/small, anchor=west] at (0.33,2.53) {Entity};
  \node[poa/small, anchor=west] at (1.71,2.53) {Type};
  \node[poa/small, anchor=east] at (3.41,2.53) {Response};
  \draw[draw=PoABorder, line width=\PoAFigureLineWidth] (0.33,2.40) -- (3.41,2.40);
  \node[poa/label, anchor=west] at (0.33,2.26) {RMC-612};
  \node[poa/label, anchor=west] at (1.71,2.26) {drug};
  \node[poa/label, anchor=east] at (3.41,2.26) {41\%};
  \node[poa/label, anchor=west] at (0.33,2.02) {TAK-881};
  \node[poa/label, anchor=west] at (1.71,2.02) {drug};
  \node[poa/label, anchor=east] at (3.41,2.02) {69\%};

  \draw[poa/sr card] (0,0.49) rectangle (3.75,1.49);
  \node[poa/sr title] at (0.20,1.30)
    {\PoAIcon[2.5mm]{lock-keyhole}\quad\PoASRReferenceTitle};
  \node[poa/small, anchor=west] at (0.20,.99) {\PoASRDerive};
  \node[poa/label, anchor=west] at (0.20,.67) {\PoASRReference};
  \draw[poa/arrow] (1.875,1.83) -- (1.875,1.49);
  \node[poa/sr transfer] at (2.45,1.66) {$(W,f)$};

  \draw[poa/sr card] (4.30,1.83) rectangle (10.40,3.81);
  \node[poa/sr title] at (4.50,3.63)
    {\PoAIcon[2.6mm]{file-text}\quad\PoASRPublicTitle};
  \draw[draw=PoABorder, line width=\PoAFigureLineWidth] (4.50,3.45) -- (10.20,3.45);
  \node[poa/small, anchor=base west] at (4.50,3.16) {Report $R$};
  \node[poa/sr body, anchor=base west, text width=4.30cm] at (5.90,3.16) {\PoASRReport};
  \node[poa/small, anchor=base west] at (4.50,2.56) {Question $q$};
  \node[poa/sr body, anchor=base west, text width=4.30cm] at (5.90,2.56) {\PoASRQuestion};
  \node[poa/small, anchor=base west] at (4.50,1.95) {Format};
  \node[poa/sr body, anchor=base west] at (5.90,1.95) {\PoASRFormat};
  \draw[poa/arrow] (3.75,2.96) -- (4.30,2.96);
  \node[poa/sr transfer,inner sep=0pt] at (4.025,3.19) {$T\!,\!W$};

  \draw[poa/sr card, draw=PoAAccentBorder] (11.15,0.49) rectangle (16.30,5.10);
  \begin{scope}
    \clip[rounded corners=\PoAFigureRadius] (11.15,0.49) rectangle (16.30,5.10);
    \fill[PoAAccent] (11.15,4.70) rectangle (16.30,5.10);
  \end{scope}
  \node[poa/sr title, poa/inverse text] at (11.35,4.90) {\PoASRClientTitle};
  \draw[poa/sr card, poa/accent soft] (11.35,3.48) rectangle (16.10,4.58);
  \node[poa/sr step] at (11.71,4.35) {1};
  \node[poa/sr prover title] at (11.98,4.27) {\PoASRFactsTitle};
  \node[poa/sr prover symbol] at (15.90,4.27) {$\widehat W$};
  \node[poa/sr prover body] at (11.55,4.14) {\PoASRFactsBody};

  \draw[poa/sr card, poa/accent soft] (11.35,2.22) rectangle (16.10,3.32);
  \node[poa/sr step] at (11.71,3.09) {2};
  \node[poa/sr prover title] at (11.98,3.01) {\PoASRQueryTitle};
  \node[poa/sr prover symbol] at (15.90,3.01) {$\widehat f$};
  \node[poa/sr prover body] at (11.55,2.88) {\PoASRQueryBody};

  \draw[poa/sr card, draw=PoAAccentBorder] (11.35,.66) rectangle (16.10,2.06);
  \node[poa/sr step] at (11.71,1.83) {3};
  \node[poa/sr prover title,text width=3.90cm] at (11.98,1.75) {\PoASRCompareTitle};
  \node[poa/label] at (13.725,1.44) {\PoASRCompareBody};
  \node[poa/badge, draw=none, fill=PoAAccentSoft, text=PoAFg,
    inner xsep=5pt, inner ysep=2pt, outer sep=0pt] (sr-answer) at (13.725,0.99)
    {\PoASRClientBody};
  \node[fit=(sr-answer), inner sep=1.5pt, outer sep=0pt] (sr-answer-focus) {};
  \draw[poa/sr focus]
    ([yshift=-1.2mm]sr-answer-focus.north west) -- (sr-answer-focus.north west) -- ++(1.8mm,0)
    ([xshift=-1.8mm]sr-answer-focus.north east) -- (sr-answer-focus.north east) -- ++(0,-1.2mm)
    ([yshift=1.2mm]sr-answer-focus.south east) -- (sr-answer-focus.south east) -- ++(-1.8mm,0)
    ([xshift=1.8mm]sr-answer-focus.south west) -- (sr-answer-focus.south west) -- ++(0,1.2mm);
  \draw[poa/accent arrow] (10.40,2.96) -- (10.775,2.96) -- (10.775,3.98) -- (11.35,3.98);
  \node[poa/sr transfer, text=PoAAccent] at (10.775,4.20) {$c$};
  \draw[poa/accent arrow] (sr-answer.west) -- (10.40,0.99);
  \node[poa/sr transfer, text=PoAAccent] at (10.775,1.26) {$\widehat a$};

  \draw[poa/sr card] (7.10,0.49) rectangle (10.40,1.49);
  \node[poa/sr title] at (7.30,1.30) {\PoASRNormalizeTitle};
  \node[poa/small, anchor=north west] at (7.30,1.10) {\PoASRNormalizeBody};
  \draw[poa/sr card, draw=PoAPrimary] (4.30,0.49) rectangle (6.50,1.49);
  \begin{scope}
    \clip[rounded corners=\PoAFigureRadius] (4.30,0.49) rectangle (6.50,1.49);
    \fill[PoAPrimary] (4.30,1.12) rectangle (6.50,1.49);
  \end{scope}
  \node[poa/sr title, poa/inverse text] at (4.50,1.30) {\PoASRCheckTitle};
  \node[poa/label, anchor=west] at (4.50,.97) {\PoASRCheckBody};
  \node[poa/small, anchor=west] at (4.50,.66) {\PoASRResult};
  \draw[poa/arrow] (7.10,0.99) -- (6.50,0.99);
  \draw[poa/arrow] (3.75,0.99) -- (4.30,0.99);
\end{tikzpicture}%

%% file: tables/01_access.tex
\begin{tabular*}{\columnwidth}{@{\extracolsep{\fill}}
  >{\raggedright\arraybackslash}m{.66\columnwidth}ccc@{\hspace{.3em}}}
\toprule
\textbf{Interface / examples} & \textbf{Users}\tnote{a} & \textbf{Ops.}\tnote{b} & \textbf{Signals}\tnote{c}\\
\midrule
\PoAAccessEntry{GUI interfaces}{Browser GUI; WebView~\cite{w3cwebaccess,androidwebview}}
  & \PoAClientIcons{1}{0}{1} & R/W & H/B/E\\
\addlinespace[.2ex]
\PoAAccessEntry{HTTP / API clients}{curl; SDKs~\cite{curlmanual}}
  & \PoAClientIcons{0}{1}{1} & R/W & H\\
\addlinespace[.2ex]
\PoAAccessEntry{Browser automation}{Playwright; Selenium~\cite{playwrightremote,seleniumoverview}}
  & \PoAClientIcons{0}{1}{1} & R/W & H/B/E\\
\addlinespace[.2ex]
\PoAAccessEntry{Browser extensions}{Content scripts; Claude in Chrome~\cite{chromecontentscripts,anthropicchrome}}
  & \PoAClientIcons{0}{1}{1} & R/W & H/B/E\\
\addlinespace[.2ex]
\PoAAccessEntry{Search / fetch}{\texttt{web\_search}; \texttt{web\_fetch}~\cite{anthropicwebsearch,anthropicwebfetch}}
  & \PoAClientIcons{0}{0}{1} & R & Var.\tnote{d}\\
\addlinespace[.2ex]
\PoAAccessEntry{Agent browser frameworks}{Browser Use~\cite{browseruseagents}}
  & \PoAClientIcons{0}{0}{1} & R/W & H/B/E\\
\addlinespace[.2ex]
\PoAAccessEntry{Agent desktop tools}{Computer Use~\cite{anthropiccomputeruse}}
  & \PoAClientIcons{0}{0}{1} & R/W & H/B/E\\
\addlinespace[.2ex]
\PoAAccessEntry{MCP tools}{Brave; Stripe; GitHub~\cite{bravemcp,stripe2026mcp,githubmcp}}
  & \PoAClientIcons{0}{0}{1} & R/W & Var.\tnote{d}\\
\bottomrule
\end{tabular*}

%% file: figures/04_passive.tex
\newcommand{\PoAPARecords}{Service-visible observations}%
\newcommand{\PoAPAHTTP}{HTTP requests}%
\newcommand{\PoAPAEnvironment}{Browser environment}%
\newcommand{\PoAPAPage}{Page events}%
\newcommand{\PoAPAFeatures}{Feature construction}%
\newcommand{\PoAPAVector}{Feature vector $x=\Phi(h)$}%
\newcommand{\PoAPAEstimator}{Passive estimator}%
\newcommand{\PoAPAOutput}{Passive estimate}%
\newcommand{\PoAPAProbability}{$p_A(x)\in[0,1]$}%
\newcommand{\PoAPAUnavailable}{Defer estimate}%
\newcommand{\PoAPAMissingData}{Insufficient observations}%
\newcommand{\PoAPAOnline}{ONLINE}%
\newcommand{\PoAPAOffline}{OFFLINE}%

\begin{tikzpicture}[poa/picture]
  \path[use as bounding box] (-.16,.65) rectangle (17.66,6.13);
  \tikzset{
    poa/pa title/.style={poa/title,anchor=west,outer sep=0pt},
    poa/pa subtitle/.style={poa/label,font=\PoAFigureFont\bfseries,anchor=west},
    poa/pa body/.style={poa/small,anchor=west},
    poa/pa neuron/.style={circle,draw=PoAAccentBorder,fill=PoAAccentSoft,
      line width=\PoAFigureLineWidth,minimum size=2.4mm,inner sep=0pt,outer sep=0pt},
    poa/pa connection/.style={draw=PoARing,line width=\PoAFigureLineWidth},
    poa/pa focus/.style={draw=PoAAccent,line width=\PoAFigureArrowWidth,
      line cap=rect,line join=miter,rounded corners=0pt}
  }

  \path[fill=PoAMuted,draw=none,rounded corners=5pt]
    (-.16,.65) rectangle (17.66,5.08);

  \begin{scope}[yshift=-.12cm]
  \path[poa/panel,draw=PoAPrimary] (-.16,5.35) rectangle (17.66,6.25);
  \begin{scope}
    \clip[rounded corners=\PoAFigureRadius] (-.16,5.35) rectangle (17.66,6.25);
    \fill[PoAPrimary] (-.16,5.35) rectangle (1.76,6.25);
  \end{scope}
  \node[poa/pa title,poa/inverse text] at (.02,5.80)
    {\PoAIcon[2.6mm]{workflow}\hspace{2pt}\PoAPAOffline};
  \foreach \xl/\xr in {1.96/6.18,6.56/8.80,9.18/11.03}
    \path[poa/card] (\xl,5.48) rectangle (\xr,6.12);
  \path[poa/card,poa/accent soft] (11.41,5.48) rectangle (13.51,6.12);
  \path[poa/card,draw=PoAPrimary] (13.89,5.48) rectangle (17.30,6.12);
  \node[poa/label,font=\PoAFigureFont\bfseries] at (4.07,5.94)
    {\PoAIcon[2.5mm]{database}\hspace{3pt}Labeled records $(h,y)$};
  \node[poa/small] at (4.07,5.65)
    {\PoAIcon[2.4mm]{user-round}\hspace{2pt}Human\hspace{5pt}%
     \PoAIcon[2.4mm]{bot}\hspace{2pt}Bot\hspace{5pt}%
     \PoAIcon[2.4mm]{brain-circuit}\hspace{2pt}Agent};
  \node[poa/small] at (7.68,5.94)
    {\PoAIcon[2.5mm]{sliders-horizontal}\hspace{2pt}Feature map $\Phi$};
  \node[poa/small] at (7.68,5.65) {used online};
  \node[poa/small] at (10.105,5.94)
    {\PoAIcon[2.5mm]{cpu}\hspace{2pt}Train $g_\theta$};
  \node[poa/small] at (10.105,5.65) {training split};
  \node[poa/small] at (12.46,5.94)
    {\PoAIcon[2.5mm]{gauge}\hspace{2pt}Calibrate $c$};
  \node[poa/small] at (12.46,5.65) {held-out split};
  \node[poa/small] at (15.595,5.94)
    {\PoAIcon[2.5mm]{lock-keyhole}\hspace{2pt}Freeze $\Phi,g_\theta,c$};
  \node[poa/small] at (15.595,5.65) {fixed for online use};
  \foreach \xl/\xr in {6.18/6.56,8.80/9.18,11.03/11.41,13.51/13.89}
    \draw[poa/arrow] (\xl,5.80) -- (\xr,5.80);
  \end{scope}

  \draw[poa/accent arrow] (15.595,5.36) -- (15.595,4.91) -- (6.975,4.91) -- (6.975,4.59);
  \draw[poa/accent arrow] (11.70,4.91) -- (11.70,4.59);
  \node[poa/small,text=PoAAccent,anchor=south] at (11.285,4.95) {Online deployment};

  \begin{scope}[yshift=-.76cm]
  \node[poa/pa title] at (.02,5.58)
    {\PoAIcon[2.8mm]{scan-eye}\hspace{3pt}\PoAPAOnline};

  \path[poa/card,draw=PoAPrimary] (0,1.61) rectangle (4.35,5.35);
  \begin{scope}
    \clip[rounded corners=\PoAFigureRadius] (0,1.61) rectangle (4.35,5.35);
    \fill[PoAPrimary] (0,4.95) rectangle (4.35,5.35);
  \end{scope}
  \node[poa/pa title,poa/inverse text] at (.20,5.15) {1. \PoAPARecords};

  \path[poa/card] (4.80,2.05) rectangle (9.15,5.35);
  \node[poa/pa title] at (5.00,5.15) {2. \PoAPAFeatures};
  \draw[draw=PoABorder,line width=\PoAFigureLineWidth] (5.00,4.95) -- (8.95,4.95);

  \path[poa/card,draw=PoAAccentBorder] (9.60,2.05) rectangle (13.80,5.35);
  \begin{scope}
    \clip[rounded corners=\PoAFigureRadius] (9.60,2.05) rectangle (13.80,5.35);
    \fill[PoAAccent] (9.60,4.95) rectangle (13.80,5.35);
  \end{scope}
  \node[poa/pa title,poa/inverse text] at (9.80,5.15) {3. \PoAPAEstimator};

  \path[poa/card,draw=PoAPrimary] (14.25,2.05) rectangle (17.50,5.35);
  \begin{scope}
    \clip[rounded corners=\PoAFigureRadius] (14.25,2.05) rectangle (17.50,5.35);
    \fill[PoAPrimary] (14.25,4.95) rectangle (17.50,5.35);
  \end{scope}
  \node[poa/pa title,poa/inverse text] at (14.45,5.15) {4. \PoAPAOutput};

  \path[poa/panel] (.18,3.95) rectangle (4.17,4.80);
  \node[poa/pa subtitle] at (.40,4.61)
    {\PoAIcon[2.5mm]{server}\hspace{3pt}\PoAPAHTTP};
  \node[poa/pa body] at (.40,4.36) {Method \textperiodcentered{} Resource type \textperiodcentered{} Status};
  \node[poa/pa body] at (.40,4.12) {Header order \textperiodcentered{} Bytes \textperiodcentered{} Timestamp};

  \path[poa/panel] (.18,2.98) rectangle (4.17,3.83);
  \node[poa/pa subtitle] at (.40,3.64)
    {\PoAIcon[2.5mm]{fingerprint-pattern}\hspace{3pt}\PoAPAEnvironment};
  \node[poa/pa body] at (.40,3.39) {Languages \textperiodcentered{} Time zone \textperiodcentered{} \texttt{webdriver}};
  \node[poa/pa body] at (.40,3.15) {Viewport \textperiodcentered{} Screen \textperiodcentered{} Font probes};

  \path[poa/panel] (.18,1.71) rectangle (4.17,2.86);
  \node[poa/pa subtitle] at (.40,2.67)
    {\PoAIcon[2.5mm]{activity}\hspace{3pt}\PoAPAPage};
  \node[poa/pa body] at (2.35,2.67) {Time};
  \node[poa/pa body] at (3.12,2.67) {Target};
  \draw[draw=PoABorder,line width=\PoAFigureLineWidth] (.40,2.52) -- (3.95,2.52);
  \foreach \yy/\ev/\tt/\target in {
    2.38/click/t_1/button,
    2.14/input/t_2/field,
    1.90/scroll/t_3/page} {
    \node[poa/pa body] at (.40,\yy) {\ev};
    \node[poa/pa body] at (2.35,\yy) {$\tt$};
    \node[poa/pa body] at (3.12,\yy) {\target};
  }

  \foreach \yb/\yt in {4.16/4.80,3.40/4.04,2.64/3.28}
    \path[poa/panel] (4.98,\yb) rectangle (8.97,\yt);
  \node[poa/pa subtitle] at (5.20,4.60)
    {\PoAIcon[2.5mm]{server}\hspace{3pt}Request statistics};
  \node[poa/pa body] at (5.20,4.30) {Count \textperiodcentered{} Concurrency \textperiodcentered{} Span \textperiodcentered{} Gap};
  \node[poa/pa subtitle] at (5.20,3.84)
    {\PoAIcon[2.5mm]{fingerprint-pattern}\hspace{3pt}Environment indicators};
  \node[poa/pa body] at (5.20,3.54) {Window \textperiodcentered{} Screen \textperiodcentered{} Browser flags};
  \node[poa/pa subtitle] at (5.20,3.08)
    {\PoAIcon[2.5mm]{clock-3}\hspace{3pt}Interaction and timing};
  \node[poa/pa body] at (5.20,2.78) {Counts \textperiodcentered{} Event order \textperiodcentered{} Time gaps};
  \path[poa/card,fill=PoAMuted] (4.98,2.20) rectangle (8.97,2.52);
  \node[poa/label,font=\PoAFigureFont\bfseries] at (6.975,2.36) {\PoAPAVector};

  \foreach \ya in {4.58,4.26,3.94} {
    \foreach \yb in {4.68,4.40,4.12,3.84} {
      \draw[poa/pa connection] (10.60,\ya) -- (11.46,\yb);
      \draw[poa/pa connection] (11.46,\yb) -- (12.32,\ya);
    }
    \draw[poa/pa connection] (12.32,\ya) -- (13.18,4.26);
  }
  \foreach \ii/\yy in {1/4.58,2/4.26,3/3.94} {
    \node[poa/pa neuron,fill=PoABackground] (pa-input-\ii) at (10.60,\yy) {};
    \node[poa/pa neuron] at (12.32,\yy) {};
  }
  \foreach \yy in {4.68,4.40,4.12,3.84}
    \node[poa/pa neuron] at (11.46,\yy) {};
  \node[poa/pa neuron,draw=PoAAccent] (pa-network-out) at (13.18,4.26) {};
  \draw[poa/arrow] (9.80,4.26) -- (pa-input-2.west);
  \node[poa/label,font=\PoAFigureSymbolFont] at (10.18,4.50) {$x$};
  \path[poa/card,fill=PoAMuted] (9.80,3.19) rectangle (13.60,3.56);
  \node[poa/label] at (11.70,3.375) {Raw score $s=g_\theta(x)$};
  \draw[poa/arrow] (pa-network-out.south) -- (13.18,3.56);
  \draw[poa/arrow] (11.70,3.19) -- (11.70,2.97);

  \path[poa/card,poa/accent soft] (9.80,2.20) rectangle (13.60,2.97);
  \node[poa/pa subtitle] at (10.00,2.75)
    {\PoAIcon[2.5mm]{gauge}\hspace{2pt}Apply calibration $c$};
  \node[poa/label,font=\PoAFigureSymbolFont] at (11.70,2.40) {$p_A(x)=c(s)$};

  \path[poa/card,draw=PoAAccentBorder] (14.45,3.42) rectangle (17.30,4.80);
  \node[poa/small] at (15.875,4.60) {Sufficient observations};
  \node[poa/label,font=\PoAFigureFont\bfseries] at (15.875,4.30) {Agent probability};
  \node[poa/badge,draw=none,fill=PoAAccentSoft,text=PoAFg,
    minimum width=2.40cm,minimum height=.50cm,inner sep=3pt,
    font=\PoAFigureSymbolFont,outer sep=0pt] (pa-probability) at (15.875,3.83)
    {\PoAPAProbability};
  \node[fit=(pa-probability),inner sep=1.5pt,outer sep=0pt] (pa-probability-focus) {};
  \draw[poa/pa focus]
    ([yshift=-1.2mm]pa-probability-focus.north west) -- (pa-probability-focus.north west) -- ++(1.8mm,0)
    ([xshift=-1.8mm]pa-probability-focus.north east) -- (pa-probability-focus.north east) -- ++(0,-1.2mm)
    ([yshift=1.2mm]pa-probability-focus.south east) -- (pa-probability-focus.south east) -- ++(-1.8mm,0)
    ([xshift=1.8mm]pa-probability-focus.south west) -- (pa-probability-focus.south west) -- ++(0,1.2mm);
  \draw[poa/arrow] (4.35,4.26) -- (4.80,4.26);
  \node[poa/label,font=\PoAFigureSymbolFont] at (4.575,4.50) {$h$};
  \draw[poa/arrow] (8.97,2.36) -- (9.375,2.36) -- (9.375,4.26) -- (9.80,4.26);
  \draw[poa/accent arrow] (13.60,2.40) -- (14.025,2.40) -- (14.025,3.83) -- (pa-probability.west);

  \path[poa/panel] (14.45,2.20) rectangle (17.30,3.24);
  \node[poa/small] at (15.875,3.04) {\PoAPAMissingData};
  \node[poa/label,font=\PoAFigureFont\bfseries] at (15.875,2.72)
    {\PoAIcon[2.5mm]{circle-question-mark}\hspace{3pt}\PoAPAUnavailable};
  \node[poa/small] at (15.875,2.40) {Policy-defined fallback};
  \draw[poa/dashed arrow,draw=PoAPrimary] (4.80,2.52) -- (4.575,2.52) -- (4.575,1.71) -- (15.875,1.71) -- (15.875,2.20);
  \node[poa/small,anchor=south] at (9.10,1.77) {\PoAPAMissingData};
  \end{scope}
\end{tikzpicture}%

%% file: sections/05_evaluation.tex
\section{Evaluation}
\label{sec:evaluation}

\subsection{Challenge--Response Evaluation}
\label{sec:challenge-evaluation}
\noindent\textbf{Experimental setup.}
We evaluated the proposed NLU challenge design using eight language models and two rule-based scripts for text processing on the same 100 questions.
Each challenge contains a textual report of 591--1,090 words, with a median of 819.5 words.
Appendix~\ref{app:challenge-details} provides the implementation settings.

We implement two rule-based scripts, \PoAScriptName{terminal}{Nearest Text} and \PoAScriptName{workflow}{Shallow Relation}.
Nearest Text selects answer spans using keyword overlap and text proximity.
Shallow Relation uses predefined patterns to extract entity--attribute--value relations, then retrieves attribute values or compares numbers to answer the question.
For comparison, the models receive the same report, question, and structured-answer instruction.
After extracting the first explicit \texttt{answer}, the verifier normalizes type prefixes and checks the result against the accepted-answer set.

\noindent\textbf{Challenge performance.}
Table~\ref{tab:blind-outcomes} shows that the eight models achieve 85--100\% accuracy, compared with 21\% for \PoAScriptName{terminal}{Nearest Text} and 4\% for \PoAScriptName{workflow}{Shallow Relation}.
Overall, the models answered 772 of the 800 assigned questions correctly; the remaining 28 responses (3.5\%) include incorrect answers and malformed output.
The scripts' successes show that the generation process can yield instances vulnerable to simple solvers.
A verifier could introduce a screening procedure to filter out weak instances before use, similar to candidate screening during key generation~\cite{nist2023fips1865}.

\begin{table}[t]
  \PoAFloatStyle
  \setlength{\tabcolsep}{2pt}
  \begin{threeparttable}
    \caption{NLU challenge performance and resource use.
      Each client is assigned the same 100 questions.}
    \label{tab:blind-outcomes}
    \label{tab:challenge-resources}
    \input{tables/02_challenges.tex}
    \begin{tablenotes}[normal]
      \footnotesize
      \raggedright
      \item[a] \textbf{Time:} Median observed solving time (model response or script execution time).
      \item[b] \textbf{Tokens:} Mean input and output tokens per challenge response.
    \end{tablenotes}
  \end{threeparttable}
\end{table}

\begin{figure}[t]
  \centering
  \input{figures/05_models.tex}
  \begin{minipage}[t]{.054\linewidth}
    \centering
    \PoAOverheadModelKey
  \end{minipage}%
  \begin{subfigure}[t]{.462\linewidth}
    \centering
    \includegraphics[width=\linewidth]{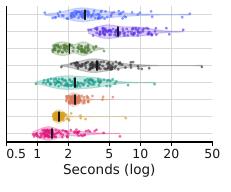}
    \caption{Response time.}
    \label{fig:challenge-client-time}
  \end{subfigure}\hfill
  \begin{subfigure}[t]{.462\linewidth}
    \centering
    \includegraphics[width=\linewidth]{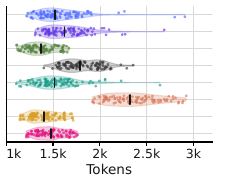}%
    \makebox[0pt][l]{%
      \hspace*{-2.164502\linewidth}%
      \raisebox{-5pt}[0pt][0pt]{%
        \includegraphics[width=2.164502\linewidth]{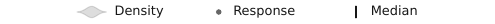}}%
    }%
    \caption{Token consumption.}
    \label{fig:challenge-token-usage}
  \end{subfigure}
  \caption{Model verification overhead.
    Response time uses a logarithmic axis and tokens use a linear axis.}
  \Description{Two panels show response times and total tokens in table order, with model logos on the left.
    Selected responses include failures; shaded widths show estimated density, points represent individual responses, and bars mark medians.}
  \label{fig:challenge-overhead}
\end{figure}

\noindent\textbf{Verification overhead.}
Next we examined the overhead introduced by this verification process.
On the client side, Figure~\ref{fig:challenge-overhead} shows the time and tokens spent answering challenges, with median response times of 1.4--6.1 seconds across the eight models.
Verification will surely incur additional token costs and processing time for agents, just as CAPTCHAs cost humans time and attention.

On the server side, Table~\ref{tab:challenge-server} reports a combined median of 327.6 microseconds for challenge assembly and response verification.
With templates prepared offline, the service only needs to assemble each challenge and check the response at runtime, keeping the overhead low for concurrent verification at scale.

\begin{table}[t]
  \PoAFloatStyle
  \setlength{\tabcolsep}{2pt}
  \begin{threeparttable}
    \caption{Server verification overhead.
      Median and P95 per-call costs, with median serialized sizes of challenges and private verification states.}
    \label{tab:challenge-server}
    \input{tables/03_overhead.tex}
    \begin{tablenotes}[normal]
      \footnotesize
      \raggedright
      \item[a] \textbf{Sizes:} Compact UTF-8 JSON; private state includes its template.
    \end{tablenotes}
  \end{threeparttable}
\end{table}

\subsection{Passive Analysis}
\label{sec:ui-evaluation}
\label{sec:http-evaluation}
\noindent\textbf{Experimental setup.}
To collect behavioral data, we used the WebArena shopping website~\cite{zhou2024webarena}, with three custom tasks: \PoATaskName{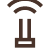}{Wi-Fi adapters}, \PoATaskName{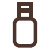}{USB drive}, and \PoATaskName{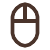}{wireless mice}.
The Wi-Fi task uses fixed instructions, while the USB and mouse tasks allow open-ended product selection within attribute and budget constraints.
We recruited twelve participants from our institute with institutional review board (IRB) approval and selected three Bot and three Agent implementations for the main comparison, as summarized in Table~\ref{tab:passive-users}.
Human participants completed the tasks manually using browsers on their own devices.
Bots followed predefined rules, while Agents used language models to interpret observations and select actions.
The adversarial settings include \textit{bots posing as humans} using native mouse and keyboard input, and \textit{bots posing as agents} using Browser Use or Playwright MCP.

\label{sec:passive-human}
During browsing, we collected HTTP metadata, browser characteristics, and page events, with pointer movement and scrolling recorded.
Finally, we retained a dataset of 720 visits: 72 Human visits (12 participants × 3 tasks × 2 rounds) and 648 automatic visits (432 Bot and 216 Agent, with 24 visits per task for each configuration). 
Appendix~\ref{app:score-details} provides detailed experimental settings.

\noindent\textbf{HTTP behavior and client fingerprints.}
Table~\ref{tab:passive-fingerprints} summarizes service-visible HTTP behavior and client fingerprints.
To characterize the networking software, we also examine JA4 and JA4H$_{ab}$ from the JA4+ suite, which provide TLS and HTTP fingerprints, respectively~\cite{foxioja4}.
From the HTTP perspective, Bot visits have a shorter median request span (4.37 s) than Human and Agent visits (39.42 and 43.55 s), while Agent visits have a longer median request gap (3220.34 ms) than Human and Bot visits (586.47 and 395.15 ms).
However, we believe these timing patterns are unreliable indicators because they are easy to alter.
From the fingerprinting perspective, Bot and Agent configurations using the same tools share identical JA4 and JA4H$_{ab}$ values, while ordinary Chrome also matches several automatic configurations, motivating us to further examine client interaction behavior for distinction.

\begin{table}[t]
  \PoAFloatStyle
  \setlength{\tabcolsep}{2pt}
  \renewcommand{\arraystretch}{1.02}
  \begin{threeparttable}
    \caption{Client configurations.
      Implementations and observation inputs for the main comparison and adversarial settings.}
    \label{tab:passive-users}
    \input{tables/04_clients.tex}
    \begin{tablenotes}[normal]
      \footnotesize
      \raggedright
      \item[a] \textbf{AX tree:} Accessibility tree, a textual description of page elements.
    \end{tablenotes}
  \end{threeparttable}
\end{table}

\begin{table}[t]
  \PoAFloatStyle
  \setlength{\tabcolsep}{2pt}
  \begin{threeparttable}
    \caption{HTTP behavior and client fingerprints.
      Median HTTP measurements and fingerprint groups across clients.}
    \label{tab:passive-fingerprints}
    \input{tables/05_fingerprints.tex}
    \begin{tablenotes}[normal]
      \footnotesize
      \raggedright
      \item[a] \textbf{Span:} Time between the first and last request starts within a visit.
      \item[b] \textbf{Gap:} Median interval between consecutive request starts within a visit.
    \end{tablenotes}
  \end{threeparttable}
\end{table}

\begin{figure}[t]
  \centering
  \captionsetup[subfigure]{
    skip=2pt,
    font={small,color=PoAPlotText},
    labelfont=bf,
    textfont=bf,
    justification=centering
  }
  \begin{minipage}[t]{.1181818\linewidth}
    \centering
    \includegraphics[width=\linewidth]{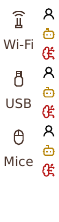}
  \end{minipage}%
  \begin{subfigure}[t]{.2939394\linewidth}
    \centering
    \includegraphics[width=\linewidth]{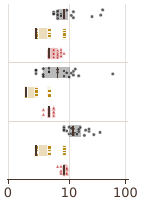}
    \caption{Clicks.}
    \label{fig:passive-clicks}
  \end{subfigure}%
  \begin{subfigure}[t]{.2939394\linewidth}
    \centering
    \includegraphics[width=\linewidth]{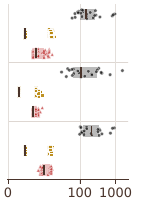}
    \caption{Pointer moves.}
    \label{fig:passive-pointer}
  \end{subfigure}%
  \begin{subfigure}[t]{.2939394\linewidth}
    \centering
    \includegraphics[width=\linewidth]{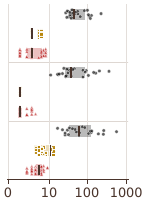}%
    \makebox[0pt][l]{%
      \hspace*{-3.402062\linewidth}%
      \raisebox{-2pt}[0pt][0pt]{%
        \includegraphics[width=3.402062\linewidth]{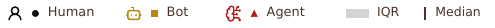}}%
    }%
    \caption{Scrolls.}
    \label{fig:passive-scroll}
  \end{subfigure}
  \caption{Interaction patterns on three shopping tasks. 
  Bars and ticks mark interquartile ranges and medians. Axes use $\log(1+x)$ scaling; points may overlap.}
  \Description{Three horizontal panels compare clicks, pointer movements, and scrolls on Wi-Fi-adapter, USB-drive, and wireless-mouse shopping tasks, with shared task and client icons.}
  \label{fig:passive-human}
\end{figure}

\noindent\textbf{Observed interaction patterns.}
\label{sec:passive-overlap}
\label{sec:passive-modality}
\label{sec:passive-extension}
Figure~\ref{fig:passive-human} compares page interactions across Humans and interaction-capable Bot and Agent implementations.
From the results, we see that Human visits generally contain more pointer movements and scrolls.
On the same tasks, even though clients may share the same interfaces, they exhibit different patterns of interaction.
Across repeated visits, Humans show the most diverse event-count combinations, while Bots are generally more repetitive, and Agents often fall between them.

\noindent\textbf{Features and model comparison.}
We extract 40 service-visible features from each visit $h$:
\[
  x=\Phi(h)=(x_{\mathrm B},x_{\mathrm A},x_{\mathrm I},x_{\mathrm S},x_{\mathrm T}),
\]
where B/A/I/S/T denote Browser, Activity, Input, Sequence, and Timing, containing 12, 5, 9, 7, and 7 features, respectively.
We specify their components in Table~\ref{tab:passive-feature-groups} (Appendix~\ref{app:score-details}).

We compare logistic regression (\textit{LR}), random forest (\textit{RF}), \textit{RBF-SVM}~\cite{cortes1995support}, and \textit{XGBoost}~\cite{chen2016xgboost} using nested cross-validation.
Each outer fold trains on two tasks and tests on the remaining task.
Human data are split by participant, with P01--P08 used for training and P09--P12 for testing; both rounds of each participant follow this assignment.
Within each outer training set, two inner folds select parameters separately for each model and feature setting.
Each task contributes 200 test visits, giving 600 test visits across the three tasks: 24 Human, 360 Bot, and 216 Agent visits.

\begin{table}[t]
  \PoAFloatStyle
  \setlength{\tabcolsep}{0pt}
  \renewcommand{\arraystretch}{1.06}
  \begin{threeparttable}
    \caption{Passive client classification.
      Rows indicate true classes\tnote{a}; columns indicate predictions, and red cells mark errors.
      Task-level macro-F1 scores\tnote{b} appear below each matrix.}
    \label{tab:passive-classification}
    \input{tables/06_classification.tex}
    \begin{tablenotes}[normal]
      \footnotesize
      \raggedright
      \item[a] \textbf{Classes:} H = Human (24), B = Bot (360), A = Agent (216).
      \item[b] \textbf{Tasks:} W = Wi-Fi adapters, U = USB drive, M = wireless mice.
    \end{tablenotes}
  \end{threeparttable}
\end{table}

\begin{figure}[t]
  \centering
  \captionsetup[subfigure]{skip=1pt}
  \begin{subfigure}[t]{.495\linewidth}
    \centering
    \includegraphics[width=\linewidth]{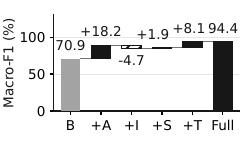}
    \caption{\textit{LR}.}
    \label{fig:passive-cumulative-lr}
  \end{subfigure}\hfill%
  \begin{subfigure}[t]{.495\linewidth}
    \centering
    \includegraphics[width=\linewidth]{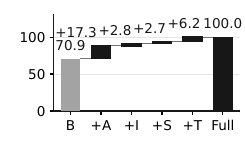}
    \caption{\textit{RF}.}
    \label{fig:passive-cumulative}
  \end{subfigure}
  \par
  \begin{subfigure}[t]{.495\linewidth}
    \centering
    \includegraphics[width=\linewidth]{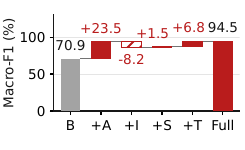}
    \caption{\textit{RBF-SVM}.}
    \label{fig:passive-cumulative-svm}
  \end{subfigure}\hfill%
  \begin{subfigure}[t]{.495\linewidth}
    \centering
    \includegraphics[width=\linewidth]{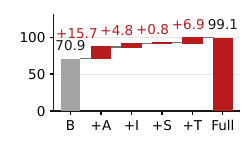}
    \caption{\textit{XGBoost}.}
    \label{fig:passive-cumulative-xgb}
  \end{subfigure}
  \caption{Cumulative feature ablation.
    B/A/I/S/T denote Browser, Activity, Input, Sequence, and Timing.
    Floating-bar labels show macro-F1 changes (pp).}
  \Description{Four compact waterfalls compare cumulative additions of Browser, Activity, Input, Sequence, and Timing features for LR, RF, RBF-SVM, and XGBoost, retaining positive, negative, and zero changes.}
  \label{fig:passive-ablation}
\end{figure}

Table~\ref{tab:passive-classification} shows that \textit{RF} achieves the highest macro-F1 (1.0000), followed by \textit{XGBoost} (0.9913), correctly labeling 600 and 593 of the 600 test visits, respectively.
All four models correctly identify the 24 Human visits, with errors confined to Bot--Agent confusions.
\textit{LR} and \textit{RBF-SVM} label 47 and 46 Bot visits as Agents, respectively, whereas \textit{XGBoost} labels 7 Agent visits as Bots.
Figure~\ref{fig:passive-ablation} shows the effects of cumulatively adding feature groups.
Activity and Timing provide the largest gains across all four models (15.7--23.5 and 6.2--8.1~pp, respectively), while Input has mixed effects and Sequence adds smaller improvements.
These results suggest the feasibility of distinguishing Humans, Bots, and Agents using passive analysis of service-visible interaction patterns.
However, given the limited scale of our dataset, we believe a larger and more comprehensive study would be a valuable direction for future work.

%% file: tables/02_challenges.tex
\begin{tabular*}{\columnwidth}{@{\extracolsep{\fill}}lrrrr@{}}
  \toprule
  \textbf{Model} & \textbf{Acc. (\%)} & \textbf{Time (s)}\tnote{a} & \textbf{Input}\tnote{b} & \textbf{Output}\tnote{b} \\
  \midrule
  \PoAModelName{\includegraphics[width=\PoAModelLogoSize]{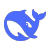}}{\texttt{DeepSeek-V4.1-Flash}} & 100 & 2.918 & 1380.9 & 200.4 \\
  \PoAModelName{\includegraphics[width=\PoAModelLogoSize]{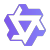}}{\texttt{Qwen3.8-Flash}} & 100 & 6.115 & 1413.2 & 238.9 \\
  \PoAModelName{\includegraphics[width=\PoAModelLogoSize]{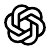}}{\texttt{GPT-5.6-Luna}} & 99 & 2.062 & 1332.5 & 42.5 \\
  \PoAModelName{\includegraphics[width=\PoAModelLogoSize]{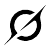}}{\texttt{Grok-4.3}} & 99 & 3.784 & 1509.4 & 307.9 \\
  \PoAModelName{\includegraphics[width=\PoAModelLogoSize]{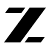}}{\texttt{GLM-5.3-Flash}} & 98 & 2.309 & 1335.0 & 201.6 \\
  \PoAModelName{\includegraphics[width=\PoAModelLogoSize]{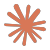}}{\texttt{Claude-Haiku-5.5}} & 96 & 2.317 & 2287.4 & 65.7 \\
  \PoAModelName{\includegraphics[width=\PoAModelLogoSize]{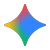}}{\texttt{Gemini-3.5-Flash-Lite}} & 95 & 1.625 & 1390.8 & 13.4 \\
  \PoAModelName{\includegraphics[width=\PoAModelLogoSize]{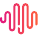}}{\texttt{MiniMax-M3.1-Flash}} & 85 & 1.400 & 1463.7 & 8.9 \\
  \midrule
  \PoAModelName{\includegraphics[width=\PoAModelLogoSize]{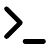}}{Nearest Text} & 21 & 0.00297 & --- & --- \\
  \PoAModelName{\includegraphics[width=\PoAModelLogoSize]{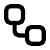}}{Shallow Relation} & 4 & 0.00189 & --- & --- \\
  \bottomrule
\end{tabular*}

%% file: tables/03_overhead.tex
\begin{tabular*}{\columnwidth}{@{\extracolsep{\fill}}lrrrrrr@{}}
  \toprule
  \multirow{2}{*}{\textbf{Phase}} & \multicolumn{2}{c}{\textbf{Wall ($\mu$s)}} & \multicolumn{2}{c}{\textbf{CPU ($\mu$s)}} & \multicolumn{2}{c}{\textbf{Instr. ($10^6$)}} \\
  \cmidrule(lr){2-3}\cmidrule(lr){4-5}\cmidrule(l){6-7}
   & \textbf{Median} & \textbf{P95} & \textbf{Median} & \textbf{P95} & \textbf{Median} & \textbf{P95} \\
  \midrule
  Assembly & 324.3 & 373.2 & 323.2 & 371.1 & 6.546 & 7.946 \\
  Verification & 3.48 & 3.91 & 3.45 & 3.89 & 0.0788 & 0.0875 \\
  Total & 327.6 & 376.6 & 326.5 & 374.6 & 6.626 & 8.024 \\
  \midrule
  \multicolumn{7}{@{}l@{}}{Challenge / private state: 6.21 / 10.06\,kB\tnote{a}} \\
  \bottomrule
\end{tabular*}

%% file: figures/05_models.tex
\newcommand{\PoAOverheadModelKey}{%
\includegraphics[width=\linewidth]{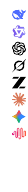}%
}

%% file: tables/04_clients.tex
\begin{tabular*}{\columnwidth}{@{\extracolsep{\fill}}lll@{}}
\toprule
\textbf{User} & \textbf{Implementation} & \textbf{Input}\tnote{a} \\
\midrule
\PoAClientIcons{1}{0}{0}\hspace{\PoAModelLogoGap}Human & Participant browser & Rendered pages \\
\addlinespace[.35ex]
\multirow{3}{*}{\PoAClientIcons{0}{1}{0}\hspace{\PoAModelLogoGap}Bot}
 & Python Requests & HTML / forms \\
 & Playwright Python & DOM \\
 & Selenium WebDriver & DOM \\
\addlinespace[.35ex]
\multirow{3}{*}{\PoAClientIcons{0}{0}{1}\hspace{\PoAModelLogoGap}Agent}
 & Browser Use & DOM \\
 & Playwright MCP & AX tree \\
 & Codex Computer Use & Screenshots \\
\midrule
\multicolumn{3}{@{}l}{\textbf{Adversarial settings}} \\
\PoAClientIcons{0}{1}{0}\hspace{.25em}poses as\hspace{.25em}\PoAClientIcons{1}{0}{0} & Native input & DOM \\
\multirow{2}{*}{\PoAClientIcons{0}{1}{0}\hspace{.25em}poses as\hspace{.25em}\PoAClientIcons{0}{0}{1}}
 & Browser Use & DOM \\
 & Playwright MCP & DOM + AX tree \\
\bottomrule
\end{tabular*}

%% file: tables/05_fingerprints.tex
\begin{tabular*}{\columnwidth}{@{\extracolsep{\fill}}lrrr@{}}
\toprule
\textbf{Attribute} & \textbf{Human (72)} & \textbf{Bot (432)} & \textbf{Agent (216)} \\
\midrule
HTTP requests & 14 & 9 & 10.5 \\
Peak concurrency & 1 & 2 & 1 \\
Request span (s)\tnote{a} & 39.42 & 4.37 & 43.55 \\
Request gap (ms)\tnote{b} & 586.47 & 395.15 & 3220.34 \\
\midrule
JA4 & T2, T3 & T1, T2 & T2 \\
JA4H$_{ab}$ & H2, H4 & H1, H2, H3 & H2, H3 \\
\bottomrule
\end{tabular*}

%% file: tables/06_classification.tex
\begin{tabular}{@{}w{l}{.18\columnwidth}*{3}{w{c}{.10333333\columnwidth}}@{\hspace{.02\columnwidth}}w{l}{.18\columnwidth}*{3}{w{c}{.10333333\columnwidth}}@{}}
\toprule
\multicolumn{4}{@{}l}{\makebox[.49\columnwidth][l]{\textbf{\textit{LR}}\hfill \textbf{Macro-F1:} 0.9440}} & \multicolumn{4}{l@{}}{\makebox[.49\columnwidth][l]{\textbf{\textit{RF}}\hfill \textbf{Macro-F1:} \textbf{1.0000}}} \\
\textbf{True$\backslash$Pred.} & \textbf{H} & \textbf{B} & \textbf{A} & \textbf{True$\backslash$Pred.} & \textbf{H} & \textbf{B} & \textbf{A} \\
\midrule
\textbf{H} (24) & \cellcolor{PoAMuted}24 & \textcolor{PoAMutedFg}{0} & \textcolor{PoAMutedFg}{0} & \textbf{H} (24) & \cellcolor{PoAMuted}24 & \textcolor{PoAMutedFg}{0} & \textcolor{PoAMutedFg}{0} \\
\textbf{B} (360) & \textcolor{PoAMutedFg}{0} & \cellcolor{PoAMuted}313 & \cellcolor{PoAAccentSoft}\textcolor{PoAAccent}{\textbf{47}} & \textbf{B} (360) & \textcolor{PoAMutedFg}{0} & \cellcolor{PoAMuted}360 & \textcolor{PoAMutedFg}{0} \\
\textbf{A} (216) & \textcolor{PoAMutedFg}{0} & \textcolor{PoAMutedFg}{0} & \cellcolor{PoAMuted}216 & \textbf{A} (216) & \textcolor{PoAMutedFg}{0} & \textcolor{PoAMutedFg}{0} & \cellcolor{PoAMuted}216 \\
\addlinespace[3pt]
\multicolumn{4}{@{}l}{\makebox[.49\columnwidth][l]{\textbf{Macro-F1:}\hfill W~1.000\hfill U~1.000\hfill M~0.837}} & \multicolumn{4}{l@{}}{\makebox[.49\columnwidth][l]{\textbf{Macro-F1:}\hfill W~1.000\hfill U~1.000\hfill M~1.000}} \\
\midrule
\multicolumn{4}{@{}l}{\makebox[.49\columnwidth][l]{\textbf{\textit{RBF-SVM}}\hfill \textbf{Macro-F1:} 0.9452}} & \multicolumn{4}{l@{}}{\makebox[.49\columnwidth][l]{\textbf{\textit{XGBoost}}\hfill \textbf{Macro-F1:} 0.9913}} \\
\textbf{True$\backslash$Pred.} & \textbf{H} & \textbf{B} & \textbf{A} & \textbf{True$\backslash$Pred.} & \textbf{H} & \textbf{B} & \textbf{A} \\
\midrule
\textbf{H} (24) & \cellcolor{PoAMuted}24 & \textcolor{PoAMutedFg}{0} & \textcolor{PoAMutedFg}{0} & \textbf{H} (24) & \cellcolor{PoAMuted}24 & \textcolor{PoAMutedFg}{0} & \textcolor{PoAMutedFg}{0} \\
\textbf{B} (360) & \textcolor{PoAMutedFg}{0} & \cellcolor{PoAMuted}314 & \cellcolor{PoAAccentSoft}\textcolor{PoAAccent}{\textbf{46}} & \textbf{B} (360) & \textcolor{PoAMutedFg}{0} & \cellcolor{PoAMuted}360 & \textcolor{PoAMutedFg}{0} \\
\textbf{A} (216) & \textcolor{PoAMutedFg}{0} & \textcolor{PoAMutedFg}{0} & \cellcolor{PoAMuted}216 & \textbf{A} (216) & \textcolor{PoAMutedFg}{0} & \cellcolor{PoAAccentSoft}\textcolor{PoAAccent}{\textbf{7}} & \cellcolor{PoAMuted}209 \\
\addlinespace[3pt]
\multicolumn{4}{@{}l}{\makebox[.49\columnwidth][l]{\textbf{Macro-F1:}\hfill W~1.000\hfill U~1.000\hfill M~0.840}} & \multicolumn{4}{l@{}}{\makebox[.49\columnwidth][l]{\textbf{Macro-F1:}\hfill W~1.000\hfill U~0.974\hfill M~1.000}} \\
\bottomrule
\end{tabular}

%% file: sections/06_discussion.tex
\section{Discussion}
\label{sec:discussion}

\noindent\textbf{Abuse Cost.}
An attacker may use a malicious agent or language model to solve a challenge while pursuing abusive goals.
This is analogous to malicious humans passing CAPTCHAs themselves or outsourcing them to CAPTCHA-solving services~\cite{motoyama2010recaptchas}.
Such model use meets the participation requirement and incurs inference costs.
Under the challenge assumptions, active verification could therefore raise the cost of access for clients that would otherwise rely entirely on scripts.
Its practical value depends on whether this added cost changes the economics of large-scale abuse without imposing excessive overhead on agents performing normal tasks.

\noindent\textbf{Verification Scope.}
From this economic perspective, verification scope affects the cost of sustained access.
A client may use a model to pass a challenge and then perform subsequent operations through scripts.
The inference cost is then spread across those operations, reducing the average cost per operation.
Verification frequency and permission scope therefore need to be considered together.
Frequent challenges add overhead to normal agent activity, whereas broad permissions allow more scripted operations after a single check.

\noindent\textbf{Practical Deployment.}
Passive analysis uses evidence from ordinary interactions and can reduce verification overhead during normal agent activity.
However, a script that is able to reproduce agent-like signals could also receive a high score without using a model.
A practical policy should consider when passive evidence is sufficient and when an active challenge is warranted.
Stricter and more frequent verification could help maintain agent participation during continued access, but would also increase the burden on ordinary agents.
The acceptability of this overhead could be explored in future deployments.
Like CAPTCHAs and bot detection mechanisms, \textit{Proof-of-Agentness} faces a usability challenge: the added verification requirements may not be welcomed by all users.

%% file: sections/07_conclusion.tex
\section{Conclusion}
We introduced \textit{Proof-of-Agentness} (\PoA{}) as a mechanism for verifying language model participation in an interaction and outlined two approaches to obtaining evidence: active verification and passive analysis.
These mechanisms could help Web services regulate automated traffic more selectively, admitting agents acting on users' behalf while limiting large-scale scripted abuse.

%% file: appendix/a_challenge_details.tex
\section{Active Verification Details}
\label{app:challenge-details}

\subsection{Implementation and Experimental Settings}
\noindent\textbf{Offline library construction.}
We prepare the library $\mathcal L$ offline using LLM-generated materials from 5 domains: biochemistry, clinical trials, cybersecurity, financial markets, and food safety.
Domain-specific prompts guide the LLM to generate entity names grouped by type and narrative templates with placeholders for entities, attributes, and values.
For each template, the LLM also supplies possible attribute values and questions paired with functions that compute their answers from the \textit{world state}.
We implement rule-based checks to ensure consistent entity types and placeholders and verify that each question refers to facts included in its report.
The resulting library contains 208 entities, 40 narrative templates, and 291 questions.
At runtime, the verifier constructs each challenge at minimal cost by sampling from this library and instantiating the selected template.

\noindent\textbf{Challenge evaluation.}
We evaluate the models on 100 questions generated from 35 of the library's 40 templates, with 20 questions from each domain.
Of these questions, 60 ask for an attribute value, 20 compare two entities, and 20 ask for the entity with the highest or lowest value.
All model queries go through OpenRouter~\cite{openrouter} with the same structured-answer requirement and no model-specific sampling or reasoning settings.
The verifier extracts the first submitted \texttt{answer} and normalizes Unicode, case, and whitespace without converting numbers or units.
We report accuracy and latency over all 100 assigned questions, counting incorrect, empty, and filtered responses as failures.
Each question contributes one selected response to the latency and token statistics.
During evaluation, Claude did not answer 2 food-safety questions because of safety filtering.
We count both as incorrect and retain their latency, while calculating Claude's mean token usage from the 98 responses with reported usage.

\noindent\textbf{Server overhead.}
We measure challenge assembly and answer checking on the same 100 questions using Python 3.9.6 on an Apple M5 running macOS arm64.
The saved correct answers serve as inputs to the verifier.
Each of the 3 independent Python processes loads resources and completes a warmup before timing.
Batching calls reduces the overhead of clock and hardware-counter readings.
For each question, we use the median per-call cost from the 3 runs; Table~\ref{tab:challenge-server} summarizes these 100 estimates.
To obtain combined overhead, we add each question's assembly and verification costs within the same run before taking medians.
The measurements exclude offline library preparation, resource loading, serialization, and HTTP serving.
Instruction counts include the Python interpreter and operating-system work.
Challenge and private-state sizes are measured using compact UTF-8 JSON serialization.

\subsection{Worked Challenge Example}
\label{app:worked-challenge}
We illustrate the construction with a biochemistry challenge from the evaluation set.
It asks which of two receptors has a higher IC50 value in a 624-word report.
The box below reproduces the complete report, question, and answer instruction received by the client.
Table~\ref{tab:challenge-example} shows the relevant values and how the verifier obtains and checks the answer.

\input{appendix/challenge_example_public.tex}

\noindent\begin{minipage}{\columnwidth}
  \captionsetup{type=table,hypcap=false}
  \PoAFloatStyle
  \setlength{\tabcolsep}{2pt}
  \begin{threeparttable}
    \caption{Example verification.
      The verifier compares two IC50 values to obtain the correct answer and check the client's response.}
    \label{tab:challenge-example}
    \input{tables/07_example.tex}
  \end{threeparttable}
\end{minipage}\par\medskip 

The correct answer is \texttt{CXCR4}, whose IC50 value is 8.35\,uM, compared with 0.14\,uM for \texttt{GABA-A}.
The verifier reads these values directly from its private \textit{world state}, whereas the client receives them only as part of the natural-language report.

%% file: appendix/challenge_example_public.tex
\begin{poachallenge}
\textbf{Report.}\par
ASSAY SCREENING PANEL SUMMARY\par
Panel reference: BIO-SCR-6; plate map revision 4.2; acquisition window 18 min per plate. The following narrative integrates inhibition curves, receptor displacement traces, and enzyme turnover checks for 6 entries. Buffer was 50 mM HEPES with 150 mM NaCl, pH 7.4, and the initial robotic dispense log also recorded fixed controls of 0.5\% DMSO, 2.0 mM MgCl2, and 37 C incubation. Compound Erlotinib was logged with replicate CV 16.3\% in the first potency table, while compound Metformin was entered nearby with replicate CV 12.8\%. No preference language is used here; these are simply panel entries.\par
The preliminary readout appeared to place the receptor curves in a narrow band. Receptor GABA-A showed Hill slope 1.58, and receptor CXCR4 showed Hill slope 0.74, with a fixed reference tracer displacement of 68\% at 100 nM. On that basis, the run captain initially wrote that receptor behavior looked essentially interchangeable. That early sentence was later softened after the kinetic annotations and replicate summaries were reconciled across the document, because slope, dispersion, and assigned mode did not remain parallel.\par
For the enzyme block, enzyme LDH-\ensuremath{\gamma} had IC50 0.72 uM after blank subtraction, and enzyme PFK-1 had IC50 64.5 uM. Their turnover checks used 12 substrate points, with a fixed calibration standard at 40.0 nmol/min/mg and a carryover flag below 0.3\%. Enzyme LDH-\ensuremath{\gamma} was separately assigned binding mode noncompetitive; enzyme PFK-1 was assigned binding mode uncompetitive. These annotations refer to the fitted inhibition pattern, not to suitability or priority.\par
Compound records were revisited after the enzyme plates were thawed a second time. Compound Erlotinib then had Hill slope 1.03 and IC50 0.31 uM; compound Metformin had Hill slope 1.76 and IC50 4.90 uM. A fixed distractor value of 91\% activity recovery appeared in the washout column for a pooled control and should not be confused with any sampled attribute. The mode calls were kept in a separate column: Erlotinib carried binding mode covalent reversible, while Metformin carried binding mode partial agonist.\par
Receptor annotations were then cross-checked against the same reporting vocabulary used for the enzyme section. Receptor GABA-A carried replicate CV 9.4\% and IC50 0.14 uM. Receptor CXCR4 carried replicate CV 25.5\% and IC50 8.35 uM. The fixed scintillation background was 145 cpm, and the ligand concentration was held at 7.5 nM. Receptor GABA-A was annotated as competitive for binding mode; receptor CXCR4 was annotated as inverse agonist for the same label.\par
The earlier impression of receptor interchangeability was not retained in the final notes. The difference was not attributed to a single number, because the final read used the full set of replicate CV, slope, dispersion, and categorical mode calls. In the same reconciliation pass, enzyme LDH-\ensuremath{\gamma} was also listed with replicate CV 5.6\% and Hill slope 0.96; enzyme PFK-1 was listed with replicate CV 1.4\% and Hill slope 0.62. A fixed note of 24 h storage at 4 C applied to all plates and is not an entity-specific measurement.\par
A final pass scattered the compound and receptor entries back into the same spreadsheet view so that no row looked privileged. Compound Erlotinib was mentioned again beside a pooled Z-prime of 0.72, and compound Metformin beside a pooled Z-prime of 0.69; neither number is the sampled IC50. Receptor GABA-A and receptor CXCR4 were also repeated in the reviewer margin to remind readers that categorical calls and quantitative values both contributed to the readout.\par
The closing assessment states only that all 6 entries were reviewed after curve fitting, replicate inspection, and nomenclature harmonization. Enzyme LDH-\ensuremath{\gamma}, enzyme PFK-1, compound Erlotinib, compound Metformin, receptor GABA-A, and receptor CXCR4 remain neutral records in the panel. The preliminary receptor sentence should be treated as provisional, while the scattered final values above are the values to use for lookup, ranking, and comparison.\par
\medskip\textbf{Question.}\par
Between GABA-A and CXCR4, which had a higher IC50?\par
\medskip\textbf{Answer instruction.}\par
For extrema, compare only entities of the requested type for which this report explicitly gives the requested attribute; missing values are not zero. Return one JSON object with exactly one key, answer, whose value is a string. Put only the exact entity name for a which-question, or the reported value for a what-question, inside that string. A numeric answer may omit its unit, but must keep the number exactly as reported. Do not convert units, expand unit names, or rewrite the numeric notation. Do not add a type prefix, explanation, Markdown, or any other field.
\end{poachallenge}

%% file: tables/07_example.tex
\begin{tabular}{@{}>{\raggedright\arraybackslash}p{.22\columnwidth}>{\raggedright\arraybackslash}p{.73\columnwidth}@{}}
\toprule
\textbf{Component} & \textbf{Content} \\
\midrule
Relevant values & \texttt{GABA-A}: 0.14 uM; \texttt{CXCR4}: 8.35 uM \\
Answer rule & Select the receptor with the higher IC50 value. \\
Correct answer & \texttt{CXCR4} \\
Example response & \texttt{\{"answer":"CXCR4"\}} \\
\bottomrule
\end{tabular}

%% file: appendix/b_passive_details.tex
\begin{table*}[t]
  \PoAFloatStyle
  \setlength{\tabcolsep}{2pt}
  \renewcommand{\arraystretch}{1.04}
  \begin{threeparttable}
    \caption{Service-visible feature groups.
      We use 40 features from 5 groups for passive client classification.}
    \label{tab:passive-feature-groups}
    \input{tables/08_features.tex}
    \begin{tablenotes}[normal]
      \footnotesize
      \raggedright
      \item[a] \textbf{Visits:} Of the 720 study visits, 648 have browser observations; the remaining 72 are from the HTTP-only Bot.
    \end{tablenotes}
  \end{threeparttable}
\end{table*}

\begin{table}[ht]
  \PoAFloatStyle
  \setlength{\tabcolsep}{3pt}
  \begin{threeparttable}
    \caption{Shopping tasks.
      All tasks start from the shopping homepage and end with cart review without checkout.}
    \label{tab:passive-tasks}
    \input{tables/09_tasks.tex}
  \end{threeparttable}
\end{table}

\section{Passive Analysis Details}
\label{app:score-details}

\subsection{Experimental Settings}
We use the 3 shopping tasks summarized in Table~\ref{tab:passive-tasks}.
The Wi-Fi adapter task specifies the exact products to add to the cart, while the other 2 allow clients to select products satisfying the stated constraints.

Browser Use and MCP Agents call \texttt{openai/gpt-6.1-sol} via OpenRouter~\cite{openrouter}, with the MCP Agent controlled through OpenAI Agents SDK.
The Computer Use Agent accesses the same model family through Codex with a ChatGPT subscription and operates through screenshots and GUI actions.
For each Computer Use visit, we use a separate conversation and Chrome profile.
The native-input Bot reads public DOM information for positioning and executes macOS mouse and keyboard events.

\balance
Each visit uses an independent guest shopping session.
Automatic clients use new browser profiles or Cookie jars, while Human participants use their own browsers and may retain resource caches.
For HTTP summaries, we retain application-page and API requests, including repetitions and failures, and exclude static resources, telemetry, and grading.
The grader checks the final cart after observation closes.
The analyzed dataset retains visits when clients do not complete the shopping tasks or their carts do not satisfy the task requirements.
For 5 of the 72 Human visits, event recording was unexpectedly incomplete; we use the available observations and treat unavailable fields as missing.
We exclude the HTTP-only Requests Bot from page-based classification.

To construct features, we apply $\log(1+x)$ to HTTP measurements, event counts, spans, and gap quantiles, keeping other features on their original scale.
Timing and sequence measurements use click, input, and scroll events, with adjacency reset at document boundaries.
The collector records pointer-move and scroll events at most once every 100\,ms each.
Within each training split, we estimate preprocessing parameters and class weights, filling missing fields with the observed means without adding missingness indicators.
\textit{LR} and \textit{RBF-SVM} also use standardization fitted on that split.
The predictors exclude identifiers and proxy-side connection counts, along with model-side logs and task grades.

Within each outer training set, we alternate the 2 tasks between inner training and validation, using P01--P04 for Human training visits and P05--P08 for validation visits.
We select among 6 prespecified candidates per classifier and feature setting by mean inner-validation macro-F1, breaking ties by candidate order, and then refit the selected classifier on all outer training data.
To supplement the interaction records, we measure TLS/HTTP fingerprints before proxy rewriting using 33 separate HTTPS probes: 3 per automatic configuration and 3 each for ordinary Chrome and Safari.
In Table~\ref{tab:passive-fingerprints}, these reference browsers supply the Human fingerprints, while the 72 participant visits provide the Human HTTP statistics.

\subsection{Ethical Considerations}
\label{app:ethics}
We obtained institutional review board (IRB) approval for the user study and informed all participants about the study and its data collection.
We collected only task-related HTTP metadata, browser characteristics, and page events within the experimental shopping website and did not collect browsing activity outside the site or any personally identifiable information.

%% file: tables/08_features.tex
\begin{tabular*}{\textwidth}{@{\extracolsep{\fill}}lcp{.72\textwidth}@{}}
\toprule
\textbf{Feature group} & \textbf{Count} & \textbf{Measurements}\hfill\textbf{Available visits}\tnote{a}\quad \PoAClientIcons{1}{0}{0}\hspace{.25em}\textbf{Human (72)}\quad \PoAClientIcons{0}{1}{0}\hspace{.25em}\textbf{Bot (360)}\quad \PoAClientIcons{0}{0}{1}\hspace{.25em}\textbf{Agent (216)} \\
\midrule
\textbf{Browser} ($x_{\mathrm{B}}$) & 12 & Inner/outer window dimensions; total/available screen dimensions; pixel ratio, color depth, maximum touch points, and webdriver flag. \\
\addlinespace[.3ex]
\textbf{Activity} ($x_{\mathrm{A}}$) & 5 & Application-request count and peak concurrency; click (including auxiliary click), input, and scroll counts. \\
\addlinespace[.3ex]
\textbf{Input} ($x_{\mathrm{I}}$) & 9 & Pointermove count, moves per click, and fraction of clicks preceded by movement; focus, paste, and composition counts; paste, composition, and deletion fractions among inputs with reported types. \\
\addlinespace[.3ex]
\textbf{Sequence} ($x_{\mathrm{S}}$) & 7 & Runs and adjacent-repeat fraction of click/input/scroll events; click$\to$input, input$\to$click, click$\to$scroll, and scroll$\to$click fractions; normalized transition entropy. \\
\addlinespace[.3ex]
\textbf{Timing} ($x_{\mathrm{T}}$) & 7 & Application-request span and median request gap; click/input/scroll span; within-document gap median, P95, and coefficient of variation; fraction of gaps above one second. \\
\bottomrule
\end{tabular*}

%% file: tables/09_tasks.tex
\begin{tabular}{@{}>{\raggedright\arraybackslash}p{.32\columnwidth}>{\raggedright\arraybackslash}p{\dimexpr.68\columnwidth-2\tabcolsep\relax}@{}}
  \toprule
  \textbf{Task} & \textbf{Cart requirements} \\
  \midrule
  \multirow{3}{=}{\PoATaskName{wifi-adapter}{Wi-Fi adapters}} &
  One specified Foktech AC1200 adapter \\
  & and one specified EDUP dual-band adapter \\
  & with an extender antenna. \\
  \addlinespace[3pt]
  \multirow{2}{=}{\PoATaskName{usb-drive}{USB drive}} &
  One USB flash drive with at least 32\,GB \\
  & capacity and a price at most \$22. \\
  \addlinespace[3pt]
  \multirow{3}{=}{\PoATaskName{wireless-mouse}{Wireless mice}} &
  Two different wireless mice, one of each, \\
  & costing at most \$20 each and \$30 combined; \\
  & at least one must support Bluetooth. \\
  \bottomrule
\end{tabular}

%% file: references.bib
@misc{foxioja4,
  author = {{FoxIO}},
  title = {{JA4+} Network Fingerprinting},
  year = {n.d.},
  url = {https://github.com/FoxIO-LLC/ja4/tree/c6469a7b0dbf8fe963810162429fbfd9d1acc4d5},
  note = {Reference implementation accessed 2026-10-25},
}

@inproceedings{park2006robot,
  author = {KyoungSoo Park and Vivek S. Pai and Kang-Won Lee and Seraphin Calo},
  title = {Securing Web Service by Automatic Robot Detection},
  year = {2006},
  booktitle = {2006 USENIX Annual Technical Conference (USENIX ATC 06)},
  pages = {255--260},
  publisher = {USENIX Association},
  url = {https://www.usenix.org/conference/2006-usenix-annual-technical-conference/securing-web-service-automatic-robot-detection},
}

@inproceedings{vonahn2003captcha,
  author = {Luis von Ahn and Manuel Blum and Nicholas J. Hopper and John Langford},
  title = {CAPTCHA: Using Hard {AI} Problems for Security},
  year = {2003},
  booktitle = {Advances in Cryptology---EUROCRYPT 2003},
  series = {Lecture Notes in Computer Science},
  volume = {2656},
  publisher = {Springer},
  address = {Berlin, Heidelberg},
  doi = {10.1007/3-540-39200-9_18},
  pages = {294--311},
  url = {https://www.iacr.org/archive/eurocrypt2003/26560294/26560294.pdf},
}

@inproceedings{motoyama2010recaptchas,
  author = {Marti Motoyama and Kirill Levchenko and Chris Kanich and Damon McCoy and Geoffrey M. Voelker and Stefan Savage},
  title = {Re: {CAPTCHAs}---Understanding {CAPTCHA}-Solving Services in an Economic Context},
  year = {2010},
  booktitle = {19th USENIX Security Symposium (USENIX Security 10)},
  publisher = {USENIX Association},
  address = {Washington, DC},
  month = aug,
  url = {https://www.usenix.org/conference/usenixsecurity10/re-captchas%E2%80%94understanding-captcha-solving-services-economic-context},
}

@inproceedings{dwork1993pricing,
  author = {Cynthia Dwork and Moni Naor},
  title = {Pricing via Processing or Combatting Junk Mail},
  year = {1993},
  booktitle = {Advances in Cryptology---CRYPTO '92},
  series = {Lecture Notes in Computer Science},
  volume = {740},
  pages = {139--147},
  publisher = {Springer},
  address = {Berlin, Heidelberg},
  doi = {10.1007/3-540-48071-4_10},
  url = {https://doi.org/10.1007/3-540-48071-4_10},
}

@inproceedings{juels1999clientpuzzles,
  author = {Ari Juels and John Brainard},
  title = {Client Puzzles: A Cryptographic Countermeasure Against Connection Depletion Attacks},
  year = {1999},
  booktitle = {NDSS},
  publisher = {Internet Society},
  numpages = {15},
  url = {https://www.ndss-symposium.org/ndss1999/cryptographic-defense-against-connection-depletion-attacks/},
}

@techreport{nist2023fips1865,
  author = {{National Institute of Standards and Technology}},
  title = {Digital Signature Standard ({DSS})},
  year = {2023},
  month = feb,
  institution = {National Institute of Standards and Technology},
  address = {Gaithersburg, MD, USA},
  type = {Federal Information Processing Standards Publication},
  number = {186-5},
  doi = {10.6028/NIST.FIPS.186-5},
  url = {https://nvlpubs.nist.gov/nistpubs/FIPS/NIST.FIPS.186-5.pdf},
}

@inproceedings{wang2019glue,
  author = {Alex Wang and Amanpreet Singh and Julian Michael and Felix Hill and Omer Levy and Samuel R. Bowman},
  title = {{GLUE}: A Multi-Task Benchmark and Analysis Platform for Natural Language Understanding},
  year = {2019},
  booktitle = {The Seventh International Conference on Learning Representations},
  url = {https://openreview.net/forum?id=rJ4km2R5t7},
}

@inproceedings{nie2020adversarial,
  author = {Yixin Nie and Adina Williams and Emily Dinan and Mohit Bansal and Jason Weston and Douwe Kiela},
  title = {Adversarial {NLI}: A New Benchmark for Natural Language Understanding},
  year = {2020},
  booktitle = {Proceedings of the 58th Annual Meeting of the Association for Computational Linguistics},
  pages = {4885--4901},
  publisher = {Association for Computational Linguistics},
  address = {Online},
  doi = {10.18653/v1/2020.acl-main.441},
  url = {https://aclanthology.org/2020.acl-main.441/},
}

@inproceedings{mccoy2019right,
  author = {R. Thomas McCoy and Ellie Pavlick and Tal Linzen},
  title = {Right for the Wrong Reasons: Diagnosing Syntactic Heuristics in Natural Language Inference},
  year = {2019},
  booktitle = {Proceedings of the 57th Annual Meeting of the Association for Computational Linguistics},
  pages = {3428--3448},
  publisher = {Association for Computational Linguistics},
  address = {Florence, Italy},
  doi = {10.18653/v1/P19-1334},
  url = {https://aclanthology.org/P19-1334/},
}

@inproceedings{levesque2012winograd,
  author = {Hector J. Levesque and Ernest Davis and Leora Morgenstern},
  title = {The Winograd Schema Challenge},
  year = {2012},
  booktitle = {Proceedings of the Thirteenth International Conference on Principles of Knowledge Representation and Reasoning},
  pages = {552--561},
  publisher = {AAAI Press},
  address = {Menlo Park, CA, USA},
  url = {https://cdn.aaai.org/ocs/4492/4492-21843-1-PB.pdf},
}

@inproceedings{weston2016babi,
  author = {Jason Weston and Antoine Bordes and Sumit Chopra and Alexander M. Rush and Bart van Merri{\"e}nboer and Armand Joulin and Tomas Mikolov},
  title = {Towards {AI}-Complete Question Answering: A Set of Prerequisite Toy Tasks},
  year = {2016},
  booktitle = {The Fourth International Conference on Learning Representations},
  url = {https://arxiv.org/abs/1502.05698},
}

@inproceedings{ribeiro2020checklist,
  author = {Marco Tulio Ribeiro and Tongshuang Wu and Carlos Guestrin and Sameer Singh},
  title = {Beyond Accuracy: Behavioral Testing of {NLP} Models with {CheckList}},
  year = {2020},
  booktitle = {Proceedings of the 58th Annual Meeting of the Association for Computational Linguistics},
  pages = {4902--4912},
  publisher = {Association for Computational Linguistics},
  address = {Online},
  doi = {10.18653/v1/2020.acl-main.442},
  url = {https://aclanthology.org/2020.acl-main.442/},
}

@inproceedings{wei2022cot,
  author = {Jason Wei and Xuezhi Wang and Dale Schuurmans and Maarten Bosma and Brian Ichter and Fei Xia and Ed Chi and Quoc V. Le and Denny Zhou},
  title = {Chain-of-Thought Prompting Elicits Reasoning in Large Language Models},
  year = {2022},
  booktitle = {Advances in Neural Information Processing Systems},
  volume = {35},
  pages = {24824--24837},
  publisher = {Curran Associates, Inc.},
  doi = {10.52202/068431-1800},
  url = {https://proceedings.neurips.cc/paper/2022/hash/9d5609613524ecf4f15af0f7b31abca4-Abstract.html},
}

@inproceedings{yao2023react,
  author = {Shunyu Yao and Jeffrey Zhao and Dian Yu and Nan Du and Izhak Shafran and Karthik Narasimhan and Yuan Cao},
  title = {{ReAct}: Synergizing Reasoning and Acting in Language Models},
  year = {2023},
  booktitle = {The Eleventh International Conference on Learning Representations},
  url = {https://openreview.net/forum?id=WE_vluYUL-X},
}

@inproceedings{gur2024webagent,
  author = {Izzeddin Gur and Hiroki Furuta and Austin Huang and Mustafa Safdari and Yutaka Matsuo and Douglas Eck and Aleksandra Faust},
  title = {A Real-World {WebAgent} with Planning, Long Context Understanding, and Program Synthesis},
  year = {2024},
  booktitle = {The Twelfth International Conference on Learning Representations},
  pages = {52690--52717},
  url = {https://proceedings.iclr.cc/paper_files/paper/2024/hash/e91bf7dfba0477554994c6d64833e9d8-Abstract-Conference.html},
}

@inproceedings{he2024webvoyager,
  author = {Hongliang He and Wenlin Yao and Kaixin Ma and Wenhao Yu and Yong Dai and Hongming Zhang and Zhenzhong Lan and Dong Yu},
  title = {{WebVoyager}: Building an End-to-End Web Agent with Large Multimodal Models},
  year = {2024},
  booktitle = {Proceedings of the 62nd Annual Meeting of the Association for Computational Linguistics (Volume 1: Long Papers)},
  pages = {6864--6890},
  publisher = {Association for Computational Linguistics},
  doi = {10.18653/v1/2024.acl-long.371},
  url = {https://aclanthology.org/2024.acl-long.371/},
}

@inproceedings{zheng2024seeact,
  author = {Boyuan Zheng and Boyu Gou and Jihyung Kil and Huan Sun and Yu Su},
  title = {{GPT}-4{V}(ision) is a Generalist Web Agent, if Grounded},
  year = {2024},
  booktitle = {Proceedings of the 41st International Conference on Machine Learning},
  series = {Proceedings of Machine Learning Research},
  volume = {235},
  pages = {61349--61385},
  publisher = {PMLR},
  url = {https://proceedings.mlr.press/v235/zheng24e.html},
}

@inproceedings{zhou2024webarena,
  author = {Shuyan Zhou and Frank F. Xu and Hao Zhu and Xuhui Zhou and Robert Lo and Abishek Sridhar and Xianyi Cheng and Tianyue Ou and Yonatan Bisk and Daniel Fried and Uri Alon and Graham Neubig},
  title = {{WebArena}: A Realistic Web Environment for Building Autonomous Agents},
  year = {2024},
  booktitle = {The Twelfth International Conference on Learning Representations},
  pages = {15585--15606},
  url = {https://proceedings.iclr.cc/paper_files/paper/2024/hash/4410c0711e9154a7a2d26f9b3816d1ef-Abstract-Conference.html},
}

@inproceedings{deng2023mind2web,
  author = {Xiang Deng and Yu Gu and Boyuan Zheng and Shijie Chen and Sam Stevens and Boshi Wang and Huan Sun and Yu Su},
  title = {{Mind2Web}: Towards a Generalist Agent for the Web},
  year = {2023},
  booktitle = {Advances in Neural Information Processing Systems},
  volume = {36},
  pages = {28091--28114},
  publisher = {Curran Associates, Inc.},
  doi = {10.52202/075280-1220},
  url = {https://proceedings.neurips.cc/paper_files/paper/2023/hash/5950bf290a1570ea401bf98882128160-Abstract-Datasets_and_Benchmarks.html},
}

@inproceedings{drouin2024workarena,
  author = {Alexandre Drouin and Maxime Gasse and Massimo Caccia and Issam H. Laradji and Manuel {Del Verme} and Tom Marty and David Vazquez and Nicolas Chapados and Alexandre Lacoste},
  title = {{WorkArena}: How Capable are Web Agents at Solving Common Knowledge Work Tasks?},
  year = {2024},
  booktitle = {Proceedings of the 41st International Conference on Machine Learning},
  series = {Proceedings of Machine Learning Research},
  volume = {235},
  pages = {11642--11662},
  publisher = {PMLR},
  url = {https://proceedings.mlr.press/v235/drouin24a.html},
}

@inproceedings{koh2024visualwebarena,
  author = {Jing Yu Koh and Robert Lo and Lawrence Jang and Vikram Duvvur and Ming Lim and Po-Yu Huang and Graham Neubig and Shuyan Zhou and Russ Salakhutdinov and Daniel Fried},
  title = {{VisualWebArena}: Evaluating Multimodal Agents on Realistic Visual Web Tasks},
  year = {2024},
  booktitle = {Proceedings of the 62nd Annual Meeting of the Association for Computational Linguistics (Volume 1: Long Papers)},
  pages = {881--905},
  publisher = {Association for Computational Linguistics},
  doi = {10.18653/v1/2024.acl-long.50},
  url = {https://aclanthology.org/2024.acl-long.50/},
}

@inproceedings{xie2024osworld,
  author = {Tianbao Xie and Danyang Zhang and Jixuan Chen and Xiaochuan Li and Siheng Zhao and Ruisheng Cao and Toh Jing Hua and Zhoujun Cheng and Dongchan Shin and Fangyu Lei and Yitao Liu and Yiheng Xu and Shuyan Zhou and Silvio Savarese and Caiming Xiong and Victor Zhong and Tao Yu},
  title = {{OSWorld}: Benchmarking Multimodal Agents for Open-Ended Tasks in Real Computer Environments},
  year = {2024},
  booktitle = {Advances in Neural Information Processing Systems},
  volume = {37},
  pages = {52040--52094},
  publisher = {Curran Associates, Inc.},
  doi = {10.52202/079017-1650},
  url = {https://proceedings.neurips.cc/paper_files/paper/2024/hash/5d413e48f84dc61244b6be550f1cd8f5-Abstract-Datasets_and_Benchmarks_Track.html},
}

@misc{mcp2026tools,
  author = {{Model Context Protocol}},
  title = {{MCP} Specification: Tools},
  year = {2026},
  url = {https://modelcontextprotocol.io/specification/2026-07-28/server/tools},
  note = {Protocol revision 2026-07-28; accessed 2026-10-14},
}

@misc{openrouter,
  author = {{OpenRouter}},
  title = {{OpenRouter} Quickstart Guide},
  year = {n.d.},
  url = {https://openrouter.ai/docs/quickstart},
  note = {Official documentation accessed 2026-10-26},
}

@misc{anthropicwebsearch,
  author = {{Anthropic}},
  title = {Web Search Tool},
  year = {n.d.},
  url = {https://platform.claude.com/docs/en/agents-and-tools/tool-use/web-search-tool},
  note = {Official documentation accessed 2026-10-21},
}

@misc{anthropicwebfetch,
  author = {{Anthropic}},
  title = {Web Fetch Tool},
  year = {n.d.},
  url = {https://platform.claude.com/docs/en/agents-and-tools/tool-use/web-fetch-tool},
  note = {Official documentation accessed 2026-10-21},
}

@misc{anthropicchrome,
  author = {{Anthropic}},
  title = {Get Started with {Claude} in {Chrome}},
  year = {n.d.},
  url = {https://support.claude.com/en/articles/12012173-get-started-with-claude-in-chrome},
  note = {Official documentation accessed 2026-10-21},
}

@misc{browseruseactor,
  author = {{Browser Use}},
  title = {Browser Actor},
  year = {n.d.},
  url = {https://github.com/browser-use/browser-use/blob/main/browser_use/actor/README.md},
  note = {Official documentation accessed 2026-10-21},
}

@misc{anthropiccomputeruse,
  author = {{Anthropic}},
  title = {Computer Use Tool},
  year = {n.d.},
  url = {https://platform.claude.com/docs/en/agents-and-tools/tool-use/computer-use-tool},
  note = {Official documentation accessed 2026-10-21},
}

@misc{browseruseagents,
  author = {{Browser Use}},
  title = {Browser Use: Open-Source Browser Agents},
  year = {n.d.},
  url = {https://github.com/browser-use/browser-use},
  note = {Official repository accessed 2026-10-23},
}

@misc{seleniumoverview,
  author = {{Selenium}},
  title = {Selenium Overview},
  year = {n.d.},
  url = {https://www.selenium.dev/documentation/overview/},
  note = {Official documentation accessed 2026-10-23},
}

@inproceedings{sardana2024inference,
  author = {Nikhil Sardana and Jacob Portes and Sasha Doubov and Jonathan Frankle},
  title = {Beyond Chinchilla-Optimal: Accounting for Inference in Language Model Scaling Laws},
  booktitle = {Proceedings of the 41st International Conference on Machine Learning},
  series = {Proceedings of Machine Learning Research},
  volume = {235},
  pages = {43445--43460},
  year = {2024},
  publisher = {PMLR},
  url = {https://proceedings.mlr.press/v235/sardana24a.html},
}

@misc{githubmcp,
  author = {{GitHub}},
  title = {{GitHub MCP Server}},
  year = {n.d.},
  url = {https://github.com/github/github-mcp-server},
  note = {Official repository accessed 2026-10-21},
}

@misc{stripe2026mcp,
  author = {{Stripe}},
  title = {Model Context Protocol ({MCP})},
  year = {2026},
  url = {https://docs.stripe.com/mcp},
  note = {Documentation accessed 2026-10-21},
}

@misc{cloudflare2025signedagents,
  author = {Jin-Hee Lee},
  title = {The age of agents: cryptographically recognizing agent traffic},
  year = {2025},
  month = aug,
  howpublished = {Cloudflare Blog},
  url = {https://blog.cloudflare.com/signed-agents/},
  note = {Published August 28, 2025; accessed 2026-10-14},
}

@misc{cloudflare2026verifiedbots,
  author = {{Cloudflare}},
  title = {Verified bots},
  year = {2026},
  url = {https://developers.cloudflare.com/bots/concepts/bot/verified-bots/},
  note = {Updated October 8, 2026; accessed 2026-10-14},
}

@misc{cloudflare2026webbotauth,
  author = {{Cloudflare}},
  title = {Web Bot Auth},
  year = {2026},
  url = {https://developers.cloudflare.com/bots/reference/bot-verification/web-bot-auth/},
  note = {Updated October 8, 2026; accessed 2026-10-14},
}

@misc{cloudflare2026detection,
  author = {{Cloudflare}},
  title = {Bot detection engines},
  year = {2026},
  url = {https://developers.cloudflare.com/bots/concepts/bot-detection-engines/},
  note = {Documentation accessed 2026-10-14},
}

@misc{cloudflare2026botscore,
  author = {{Cloudflare}},
  title = {Bot scores},
  year = {2026},
  url = {https://developers.cloudflare.com/bots/concepts/bot-score/},
  note = {Documentation accessed 2026-10-14},
}

@misc{cloudflare2026turnstile,
  author = {{Cloudflare}},
  title = {Turnstile},
  year = {2026},
  url = {https://developers.cloudflare.com/turnstile/},
  note = {Documentation accessed 2026-10-24},
}

@misc{cloudflare2026agenticweb,
  author = {Matthew Conroy},
  title = {The Internet has a second audience},
  year = {2026},
  month = sep,
  howpublished = {Cloudflare Blog},
  url = {https://blog.cloudflare.com/agentic-web/},
  note = {Published September 30, 2026; accessed 2026-10-21},
}

@misc{cloudflare2026markdownagents,
  author = {Celso Martinho and Will Allen},
  title = {Introducing {Markdown} for Agents},
  year = {2026},
  month = feb,
  howpublished = {Cloudflare Blog},
  url = {https://blog.cloudflare.com/markdown-for-agents/},
  note = {Published February 12, 2026; accessed 2026-10-15},
}

@misc{vercel2026agentpages,
  author = {Zach Cowan and Mitul Shah},
  title = {Making agent-friendly pages with content negotiation},
  year = {2026},
  month = feb,
  howpublished = {Vercel Blog},
  url = {https://vercel.com/blog/making-agent-friendly-pages-with-content-negotiation},
  note = {Published February 3, 2026; accessed 2026-10-15},
}

@misc{bravemcp,
  author = {{Brave}},
  title = {{Brave Search MCP Server}},
  year = {n.d.},
  url = {https://github.com/brave/brave-search-mcp-server},
  note = {Official repository accessed 2026-10-21},
}

@misc{examcp,
  author = {{Exa}},
  title = {{Exa MCP}},
  year = {n.d.},
  url = {https://exa.ai/docs/get-started/exa-mcp},
  note = {Documentation accessed 2026-10-21},
}

@misc{tavilymcp,
  author = {{Tavily}},
  title = {{Tavily MCP Server}},
  year = {n.d.},
  url = {https://docs.tavily.com/documentation/mcp},
  note = {Documentation accessed 2026-10-21},
}

@misc{huggingfacemcp,
  author = {{Hugging Face}},
  title = {{Hugging Face MCP Server}},
  year = {n.d.},
  url = {https://huggingface.co/docs/hub/en/agents-mcp},
  note = {Documentation accessed 2026-10-21},
}

@misc{kagglemcp,
  author = {{Kaggle}},
  title = {{MCP Server Documentation}},
  year = {n.d.},
  url = {https://www.kaggle.com/docs/mcp},
  note = {Documentation accessed 2026-10-21},
}

@misc{wikidatamcp,
  author = {{Wikidata contributors}},
  title = {{Wikidata:MCP}},
  year = {n.d.},
  url = {https://www.wikidata.org/wiki/Wikidata:MCP},
  note = {Project documentation accessed 2026-10-21},
}

@misc{paypal2026mcp,
  author = {{PayPal}},
  title = {{MCP} server quickstart guide},
  year = {2026},
  url = {https://developer.paypal.com/ai-tools/mcp-server},
  note = {Updated June 30, 2026; accessed 2026-10-21},
}

@misc{shopifyagents,
  author = {{Shopify}},
  title = {Build commerce agents with {UCP}},
  year = {n.d.},
  url = {https://shopify.dev/docs/agents},
  note = {Documentation accessed 2026-10-21},
}

@misc{notionmcp,
  author = {{Notion}},
  title = {{Notion MCP}},
  year = {n.d.},
  url = {https://www.notion.com/help/notion-mcp},
  note = {Documentation accessed 2026-10-21},
}

@misc{atlassianmcp,
  author = {{Atlassian}},
  title = {{Atlassian MCP Server}},
  year = {n.d.},
  url = {https://atlassian.github.io/atlassian-mcp-server/},
  note = {Documentation accessed 2026-10-21},
}

@inproceedings{garzon2026agentidentity,
  author = {Sandro Rodriguez Garzon and Awid Vaziry and Enis Mert Kuzu and Dennis Enrique Gehrmann and Buse Varkan and Alexander Gaballa and Axel K{\"u}pper},
  title = {{AI} Agents with Decentralized Identifiers and Verifiable Credentials},
  year = {2026},
  booktitle = {Proceedings of the 18th International Conference on Agents and Artificial Intelligence - Volume 1: ICAART},
  pages = {252--259},
  publisher = {SciTePress},
  doi = {10.5220/0014234400004052},
  url = {https://doi.org/10.5220/0014234400004052},
}

@inproceedings{xu2026agentdid,
  author = {Minghui Xu and Xiaoyu Liu and Yihao Guo and Chunchi Liu and Yue Zhang and Xiuzhen Cheng},
  title = {{AgentDID}: Trustless Identity Authentication for {AI} Agents},
  year = {2026},
  booktitle = {2026 IEEE 46th International Conference on Distributed Computing Systems (ICDCS)},
  pages = {1180--1190},
  publisher = {IEEE},
  doi = {10.1109/2575-8411.2026.00116},
  url = {https://doi.org/10.1109/2575-8411.2026.00116},
}

@inproceedings{wang2026fpagent,
  author = {Ethan Wang and Zubair Shafiq and Yash Vekaria},
  title = {Poster: {FP-Agent}: Towards Fingerprinting {AI} Browsing Agents},
  booktitle = {Proceedings of the 2026 ACM Internet Measurement Conference},
  series = {IMC '26},
  year = {2026},
  pages = {1227--1228},
  publisher = {ACM},
  doi = {10.1145/3777912.3847600},
  url = {https://doi.org/10.1145/3777912.3847600},
}

@misc{fayolle2026llmbot,
  author = {Iliana Fayolle and Sihem Bouhenniche and Samuel P{\'{e}}lissier and Pierre Laperdrix and Cl{\'{e}}mentine Maurice and Walter Rudametkin},
  title = {On the Internet, Nobody Knows You're an {LLM} Bot: Unmasking Web Agents with Multi-Layer Fingerprinting},
  year = {2026},
  howpublished = {arXiv:2606.30119v1},
  doi = {10.48550/arXiv.2606.30119},
  url = {https://arxiv.org/abs/2606.30119v1},
}

@misc{lugoloobi2026knownactions,
  author = {William Lugoloobi and Samuelle Marro and Jabez Magomere and Joss Wright and Chris Russell},
  title = {Known By Their Actions: Fingerprinting {LLM} Browser Agents via {UI} Traces},
  year = {2026},
  howpublished = {arXiv:2605.14786v1},
  doi = {10.48550/arXiv.2605.14786},
  url = {https://arxiv.org/abs/2605.14786v1},
  note = {Accepted at NeurIPS 2026. Author implementation: \url{https://github.com/KabakaWilliam/known_actions}},
}

@inproceedings{jonker2019fingerprint,
  author = {Hugo Jonker and Benjamin Krumnow and Gabry Vlot},
  title = {Fingerprint Surface-Based Detection of Web Bot Detectors},
  booktitle = {Computer Security---ESORICS 2019},
  series = {Lecture Notes in Computer Science},
  volume = {11736},
  pages = {586--605},
  year = {2019},
  publisher = {Springer International Publishing},
  address = {Cham},
  doi = {10.1007/978-3-030-29962-0_28},
  url = {https://doi.org/10.1007/978-3-030-29962-0_28},
}

@article{iliou2021advancedbots,
  author = {Christos Iliou and Theodoros Kostoulas and Theodora Tsikrika and Vasilis Katos and Stefanos Vrochidis and Ioannis Kompatsiaris},
  title = {Detection of Advanced Web Bots by Combining Web Logs with Mouse Behavioural Biometrics},
  journal = {Digital Threats: Research and Practice},
  volume = {2},
  number = {3},
  pages = {1--26},
  year = {2021},
  publisher = {Association for Computing Machinery},
  doi = {10.1145/3447815},
  url = {https://doi.org/10.1145/3447815},
}

@misc{cloudflare2026precursor,
  author = {Marina Elmore and Benedikt Wolters},
  title = {Introducing {Precursor}: Detecting Agentic Behavior with Continuous Client-Side Signals},
  year = {2026},
  month = jul,
  howpublished = {Cloudflare Blog},
  url = {https://blog.cloudflare.com/introducing-precursor/},
  note = {Published July 13, 2026; accessed 2026-10-23},
}

@misc{curlmanual,
  author = {{curl}},
  title = {curl---How To Use},
  year = {n.d.},
  url = {https://curl.se/docs/manpage.html},
  note = {Official documentation accessed 2026-10-23},
}

@misc{androidwebview,
  author = {{Android Developers}},
  title = {Build Web Apps in {WebView}},
  year = {n.d.},
  url = {https://developer.android.com/develop/ui/views/layout/webapps/webview},
  note = {Official documentation accessed 2026-10-23},
}

@misc{w3cwebaccess,
  author = {{W3C Web Accessibility Initiative}},
  title = {Tools and Techniques},
  year = {n.d.},
  url = {https://www.w3.org/WAI/people-use-web/tools-techniques/},
  note = {Official documentation accessed 2026-10-23},
}

@misc{playwrightremote,
  author = {{Microsoft}},
  title = {{BrowserType}: Connecting to an Existing Browser},
  year = {n.d.},
  url = {https://playwright.dev/python/docs/api/class-browsertype},
  note = {Official documentation accessed 2026-10-23},
}

@misc{chromecontentscripts,
  author = {{Chrome for Developers}},
  title = {Content Scripts},
  year = {n.d.},
  url = {https://developer.chrome.com/docs/extensions/develop/concepts/content-scripts},
  note = {Official documentation accessed 2026-10-23},
}

@article{cortes1995support,
  author = {Corinna Cortes and Vladimir Vapnik},
  title = {Support-Vector Networks},
  journal = {Machine Learning},
  volume = {20},
  number = {3},
  pages = {273--297},
  year = {1995},
  doi = {10.1007/BF00994018},
}

@inproceedings{chen2016xgboost,
  author = {Tianqi Chen and Carlos Guestrin},
  title = {{XGBoost}: A Scalable Tree Boosting System},
  booktitle = {Proceedings of the 22nd ACM SIGKDD International Conference on Knowledge Discovery and Data Mining},
  pages = {785--794},
  year = {2016},
  doi = {10.1145/2939672.2939785},
}
